\documentclass[letterpaper, 11pt]{article}

\usepackage[margin=1in]{geometry}
\usepackage{times}
\usepackage{helvet}
\usepackage{courier}
\usepackage[hyphens]{url}
\usepackage{graphicx}
\usepackage{natbib}
\usepackage{caption}

\usepackage{algorithm}
\usepackage{algorithmic}

\usepackage{newfloat}
\usepackage{listings}
\DeclareCaptionStyle{ruled}{labelfont=normalfont,labelsep=colon,strut=off}
\floatstyle{ruled}
\newfloat{listing}{tb}{lst}{}
\floatname{listing}{Listing}

\usepackage{amsmath}
\usepackage{amssymb}
\usepackage{mathtools}
\usepackage{amsthm}
\usepackage{enumitem}
\usepackage{siunitx}
\usepackage{microtype}

\theoremstyle{plain}
\newtheorem{theorem}{Theorem}[section] 

\newtheorem{proposition}[theorem]{Proposition}

\theoremstyle{definition}

\theoremstyle{remark}

\usepackage{booktabs}
\usepackage{multirow}
\usepackage{tabularx}
\usepackage{array}
\usepackage{makecell}
\usepackage[dvipsnames]{xcolor}
\usepackage{colortbl}
\usepackage{hyperref}
\usepackage{subcaption} 

\usepackage[most]{tcolorbox}
\newtcbox{\eqhlR}{on line, arc=2pt, boxsep=1pt, colback=red!10, colframe=red!10}
\newtcbox{\eqhlB}{on line, arc=2pt, boxsep=1pt, colback=blue!10, colframe=blue!10}

\usepackage{hyperref}

\definecolor{UpGreen}{RGB}{34, 139, 34}
\definecolor{DownOrange}{RGB}{210, 105, 30}
\definecolor{NeutralGray}{RGB}{120, 120, 120}
\newcommand{\inc}[1]{\textcolor{UpGreen}{\scriptsize\,\raisebox{0.2ex}{$\uparrow$}\,#1}}
\newcommand{\dec}[1]{\textcolor{DownOrange}{\scriptsize\,\raisebox{0.2ex}{$\downarrow$}\,#1}}
\newcommand{\sta}[1]{\textcolor{NeutralGray}{\scriptsize\,\raisebox{0.2ex}{$-$}\,#1}}

\newcommand{\ie}{\textit{i.e., }}
\newcommand{\m}{\textsc{InS}} %
\newcommand{\q}{\textsc{InSPO}} %

\newcommand{\oursrow}{\rowcolor{blue!8}}

\DeclareMathOperator{\logit}{logit}

\title{InSPO: Unlocking Intrinsic Self-Reflection for LLM Preference Optimization}

\author{Yu Li, Tian Lan, Zhengling Qi\\
George Washington University
}
\date{}

\begin{document}
\maketitle

\begin{abstract}

Direct Preference Optimization (DPO) and its variants have become the standard for aligning Large Language Models (LLMs) due to their simplicity and offline stability. However, we identify two fundamental limitations that undermine the reliability and optimality of DPO and its equivalents.
First, the optimized policy derived by current methods lacks invariance: It changes with respect to modeling choices such as the scalarization function (e.g, logistic function in Bradley-Terry model) or the reference policy. This may yield brittle behaviors of the learned policy as an artifact of parameterization rather than a reflection of true human preference, 
while an optimal policy must satisfy the invariance property.
Second, we show that the derived policy of most existing methods  is theoretically suboptimal because it
fails to fully capitalize on the comparative information embedded in the pairwise preference data. This restricts the model’s 
ability to "compare and contrast" responses,
illuminating a unique opportunity for triggering intrinsic self-reflection. 
In this work, we propose a novel family of Intrinsic Self-reflective Preference Optimization (InSPO) methods, which address these two limitations. We first derive a globally optimal policy that conditions on both the context and the alternative response under the pairwise preference data setting, which explicitly formalize the novel notion of self-reflection. Then we theoretically demonstrate that this formulation is superior to standard DPO and RLHF targets and guarantees invariance to the choice of scalarization and reference policy. Practically, InSPO operationalizes this target as a plug-and-play enhancement for DPO-family algorithms, decoupling the alignment goal from modeling constraints without requiring complex architectural changes. Crucially, leveraging the concept of learning using privileged information, our method does not require generating an alternative response during deployment, thus incurring zero extra inference overhead, as the self-reflective mechanism is distilled into the policy during training. Comprehensive experiments demonstrate that InSPO (implemented on a range of DPO-family algorithms) achieves consistent improvements in win rates and length-controlled metrics, validating that InSPO with self-reflection leads to more robust and human-aligned LLMs.
Our Code is available at \textcolor{blue}{\href{https://github.com/Skylanding/InSPO}{https://github.com/Skylanding/InSPO}}.
\end{abstract}
\section{Introduction}
\label{sec:intro}
Large language models (LLMs) are fine-tuned after pretraining through post-training alignment. This stage typically employs two complementary families of techniques~\citep{kumar2025llm}. 
The first technique is supervised fine-tuning (SFT), which fine-tunes the model to generate the correct response given an instruction, typically by minimizing the negative log-likelihood of the target tokens. SFT effectively imparts formatting conventions, stylistic preferences, and basic task-following behaviors~\citep{dong2023abilities}. 
The second technique is Reinforcement Learning from Human Feedback (RLHF), which aligns the model by learning from human pairwise preference data \citep{ouyang2022training}. This standard pipeline operates in two stages: first, a reward model is fitted to approximate human preferences, where the Bradley-Terry (BT) model is imposed to link the human preference probability with the reward contrast between two responses via the logistic function. 
Second, the LLM is fine-tuned by maximizing the learned reward with a penalty on the Kullback–Leibler (KL) divergence between the trainable policy and the reference one. The KL penalty is used to stabilize training and prevent overfitting, which is motivated by  Proximal Policy Optimization (PPO)~\citep{schulman2017proximal}. While RLHF shows promising results, it requires iterative on-policy sampling, coordinated training of reward, and careful hyperparameter tuning. The complexity and computational overhead often complicate the deployment in practice~\citep{mei2023mac,yu2025optimizing}.

Direct Preference Optimization (DPO), in contrast, simplifies this process by removing both the explicit reward model and the online rollout phase. Given a preference pair where one response is preferred over another for a given context, DPO associates a latent preference score with log-likelihood ratios between the trainable policy and a reference model and directly maximizes the margin that favors the preferred response under a logistic loss. This equivalent reformulation preserves the KL regularization implicit in RLHF while operating purely offline on the static preference pairs. 
The success of DPO has inspired a growing family of reward-free preference optimization methods that refine various aspects of the optimization objective, such as adjusting the loss curvature \citep{azar2024general}, incorporating length normalization \citep{meng2024simpo}, or modifying the reference formulation \citep{ethayarajh2024kto}, to further improve DPO.

Despite the impressive performance of RLHF and DPO-based approaches, we show that these existing methods are subject to two fundamental limitations that undermine the reliability and optimality of the learned policy. First, the derived optimal policy lacks invariance to modeling choices including the scalarization function (e.g., the logistic function in the BT model) and the reference policy. Ideally, the aligned behavior of an LLM should reflect the ground-truth human preference distribution, independent of the mathematical formulation used to process those preferences. However, as we formally demonstrate, the putative "optimal" target in current frameworks shifts with these design choices. Consequently, methods such as DPO and RLHF often produce behavior that reflects parameterization artifacts rather than genuine human preferences.

Second, these methods are theoretically suboptimal because they fail to exploit the comparative information embedded in pairwise preference data, effectively leaving the model's inherent capacity for self-reflection untapped. In particular, current approaches cast response generation as an isolated maximization problem, disregarding the fundamentally contextual nature of human preference, which is shaped by interactions between responses rather than their standalone quality~\citep{tversky1993context}. By being oblivious to
explicitly conditioning on an alternative response during the training, existing methods limit its ability to ``compare and contrast'', a mechanism we term \textit{intrinsic self-reflection}, thereby capping the potential quality of the fine-tuned policy. These two limitations raise a fundamental question for LLM preference optimization:
\begin{tcolorbox}[enhanced,colback=cyan!4,colframe=cyan!35,boxrule=0.8pt,arc=2pt,
    left=8pt,right=8pt,top=6pt,bottom=6pt,
    attach boxed title to top left={xshift=5pt,yshift=-3pt},
    boxed title style={colback=cyan!38,arc=2pt,boxrule=0pt},
    title={\small\textbf{\textcolor{white}{A Key Question}}}]
\small\centering
\textit{How can we construct a new preference optimization framework that is both \textbf{invariant to modeling choices} and capable of \textbf{fully exploiting the comparative nature} of human preferences?}
\end{tcolorbox}

\begin{figure*}[t]
    \centering
    \includegraphics[width=1.0\columnwidth]{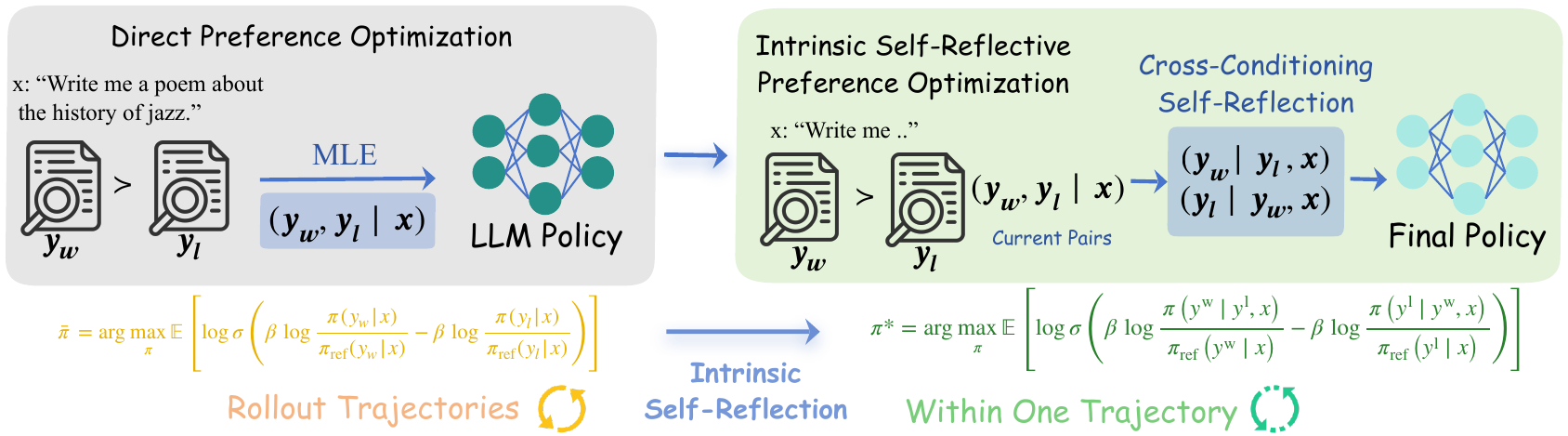}
    \caption{From pairwise preference to our proposed InSPO. Standard DPO (left) learns a suboptimal policy $\bar \pi$ from response comparisons where both the preferred and dispreferred responses are evaluated based solely on the prompt. \q~(right) unleashes intrinsic self-reflection through symmetric cross-conditioning for learning an optimal policy $\pi^\ast$: the policy generates the preferred response while seeing the dispreferred one as context, and vice versa, allowing the model to leverage alternative responses as in-context guidance for improvement. Green terms highlight the self-reflection mechanism in our new objectives.} 
    \label{fig:model}
\end{figure*}

In this work, we propose a new perspective on preference optimization that fully leverages the comparative information embedded in pairwise preference data, which is a rich signal that can be effectively capitalized using a novel notion of self-reflection.
To this end, we first derive a globally optimal policy that conditions on not only the context but also the alternative response. Then, we mathematically demonstrate that this formulation yields a target superior to standard RLHF and DPO, while enjoying invariance to the choice of scalarization and reference policy. The results enable us to develop a novel framework that unlocks Intrinsic Self-reflective Preference Optimization (InSPO) 
for existing methods using the standard pairwise preference data. Our InSPO is a plug-and-play enhancement for the DPO family via symmetric cross-conditioning that is both theoretically rigorous and computationally efficient. It effectively decouples the alignment goal from arbitrary modeling constraints and unlocks the intrinsic self-reflection capability. The alternative response is not merely as data, but as \textit{privileged information}, an important signal only available during the training phase to guide optimization \citep{vapnik2009new}. By conditioning on the alternative response during training, InSPO creates a contrastive scaffold that shapes the optimization landscape, allowing the model to learn sharper preference boundaries. Crucially, because this mechanism acts as a regularizer for the shared policy weights, the self-reflective capability is distilled into the model itself. This allows us to remove the privileged context during deployment, maintaining standard inference protocols with zero computational overhead while retaining the benefits of comparative training. Lastly, we validate the proposed InSPO methods through comprehensive experiments on a range of DPO-family algorithms, showing consistent improvements in win rates and controlled metrics across benchmarks; our analysis reveals that the proposed method induces dense reward shaping and scales effectively with model capacity, all with zero inference overhead. 
See Figure \ref{fig:model} for an overview of our method.

\section{Background and Notations}
\label{sec:prelim}

To align an LLM with human preference, most existing pipelines first collect pairwise preference data generated by a pre-trained or SFT policy $\pi_{\mathrm{ref}}$. Each sample in the preference data consists of a context $x \in \mathcal X$ generated by some distribution $\rho$, and two responses $y_w \in \mathcal Y$ and $y_\ell \in \mathcal Y$, where $\mathcal X$ and $\mathcal Y$ are the context and response spaces respectively.
After collecting such preference data, one can apply RLHF or DPO, two primary approaches, to fine-tune an LLM. 

RLHF consists of two steps. In the first step, it uses the Bradley-Terry (BT) probabilistic model to understand the human preference on $(x, y_w, y_\ell)$ and learn a reward function. Specifically, BT model assumes that
\begin{align}
\label{eq:bt}
\mathbb{P}(y_w \succ y_\ell \mid x)
&= \sigma\!\big(r(x,y_w)-r(x,y_\ell)\big),
\end{align}
where $y_w \succ y_\ell$ indicates $y_w$ is preferred to $y_\ell$, $\sigma(z)=1/(1+e^{-z})$, and $r$ is the unknown reward function to evaluate the quality of each response to the context $x$.
Then given preference data, one can implement maximum likelihood estimation (MLE) to estimate the reward function. 
In the second step, a policy optimization is executed to find a better LLM policy that maximizes the learned reward. 
For example, PPO~\citep{ouyang2022training} solves
\begin{align}\label{eqn: ppo}
    \max_{\pi} \, \mathbb{E}_{x \sim \rho, y \sim \pi}\left[r(x, y) \right] - \beta D_{\mathrm{KL}}(\pi \,\|\, \pi_{\mathrm{ref}}),
\end{align}
where $\beta >0$ is a regularization parameter that controls the strength of the KL divergence  toward the reference policy $\pi_{\mathrm{ref}}$, and the divergence $D_{\mathrm{KL}}(\pi \,\|\, \pi_{\mathrm{ref}})$ is defined as 
\[
D_{\mathrm{KL}}(\pi \,\|\, \pi_{\mathrm{ref}})
= \mathbb{E}_{x \sim \rho} \Big[ \mathrm{KL}\big(\pi(\cdot \mid x) \,\|\, \pi_{\mathrm{ref}}(\cdot \mid x)\big) \Big].
\]

As an alternative approach to RLHF, DPO shows that solving \eqref{eqn: ppo} is equivalent to modeling
\[
r(x,y_w)-r(x,y_\ell)= \beta\left(\log\frac{\pi_\theta(y_w\mid x)}{\pi_{\mathrm{ref}}(y_w\mid x)}
-\log\frac{\pi_\theta(y_\ell\mid x)}{\pi_{\mathrm{ref}}(y_\ell\mid x)}\right).
\]
Then based on the MLE loss derived under BT model \eqref{eq:bt}, DPO fine-tunes an LLM via solving
\begin{align}
\label{eq:dpo}
\max_{\pi}\,\mathbb{E}_{(x,y_w,y_\ell)\sim \mathcal D }\bigg[
\log \sigma\bigg(
\beta\log\frac{\pi(y_w\mid x)}{\pi_{\mathrm{ref}}(y_w\mid x)} 
-\beta\log\frac{\pi(y_\ell\mid x)}{\pi_{\mathrm{ref}}(y_\ell\mid x)}
\bigg)\bigg],
\end{align}
where $\mathcal D$ is the joint distribution of $(x,y_w,y_\ell)$. Here without loss of generality, we assume $y_w$ is always preferred to $y_\ell$ after rearrangement of the preference data.

As established in Proposition 1 of \cite{azar2024general}, both methods unify under a general preference optimization framework with the scalarization function $\Psi(q) = \log(q / (1-q))$. The general objective is formulated as:
\begin{align}\label{eqn: general objective}
    \max_{\pi} \;
& \mathbb{E}_{x \sim \rho}
\mathbb{E}_{y \sim \pi(\cdot\mid x), y' \sim \pi_{\mathrm{ref}}(\cdot\mid x)}
\left[
\Psi\!\big( \mathbb{P}(y \succ y' \mid x) \big)
\right].
\end{align}
Here, $\Psi: [0, 1] \rightarrow \mathbb R$ can be any non-decreasing function. Within this framework, existing methods seek an optimal policy $\bar{\pi}$ restricted to the class of context-conditioned policies $\bar \Pi = \{\pi: \mathcal X \rightarrow \mathcal Y\}$ defined as
\[
\bar{\pi} \in \mathrm{argmax}_{\pi \in \bar \Pi} \mathcal V(\pi),
\]
where the value of a policy $\mathcal V(\pi)$ is defined as
\[
\mathcal V(\pi) \triangleq \mathbb{E}_{x \sim \rho}
\mathbb{E}_{y \sim \pi(\cdot\mid x), y' \sim \pi_{\mathrm{ref}}(\cdot\mid x)}
\left[
\Psi\!\big( \mathbb{P}(y \succ y' \mid x) \big)
\right].
\]

To conclude this section, we assume that we have a dataset of $n$ pairwise preferences $\mathcal{D}_n = \{(x^{(i)}, y_w^{(i)}, y_\ell^{(i)})\}_{i=1}^n$, where in $i$-th sample, a prompt/context $x^{(i)}$ is drawn from distribution $\rho$, and two responses $(y_w^{(i)}, y_\ell^{(i)})$ are generated by the reference policy $\pi_{\mathrm{ref}}$, labeled such that $y_w^{(i)} \succ y_\ell^{(i)}$.


\section{Limitations of Existing Methods}
In this section, we identify two critical limitations of existing frameworks by investigating the properties of the restricted optimal policy $\bar{\pi}$: (i) Is $\bar{\pi}$ invariant to  the scalarization function $\Psi$ and the reference distribution $\pi_{\mathrm{ref}}$? (ii) Is $\bar{\pi}$ theoretically optimal? We demonstrate below that the answer to both questions is negative.

\subsection{Is $\bar{\pi}$ invariant to $\Psi$ and $\pi_{\mathrm{ref}}$?}
Ideally, a robust alignment framework should yield an optimal policy invariant to the choice of the scalarization function $\Psi$ and the reference policy $\pi_{\mathrm{ref}}$. This invariance property is critical for modeling robustness and disentanglement from the reference. 
Specifically, human preferences are fundamentally ordinal. The optimal policy should reflect the underlying ranking of responses, rather than being an artifact of the specific choices of mathematical transformation, \ie $\Psi$, used to process the preference probabilities or the reference policy, \ie $\pi_{\mathrm{ref}}$. In addition, the choice of $\Psi$, which is often selected for numerical stability or concavity rather than semantic relevance, should not dictate the final behavior of the model. If the optimal fine-tuned policy changes based on modeling choices, the alignment process may become brittle and inconsistent. Lastly, a principled optimization objective should decouple the learned preferences from the reference policy $\pi_{\mathrm{ref}}$. Dependence on the reference policy implies that the ``optimal" behavior is relative and transient, rather than converging toward a global optimal policy that maximizes human preference.
However, the following proposition establishes that the existing target $\bar \pi$ fails to satisfy this condition.
\begin{proposition}\label{prop: invariant}
    The form of $\bar \pi$ is not invariant to $\Psi$ and $\pi_{\mathrm{ref}}$.
\end{proposition}
In the proof of Proposition \ref{prop: invariant}, we provide counter-examples demonstrating that $\bar \pi$ shifts when either $\Psi$ or $\pi_{\mathrm{ref}}$ is varied respectively. This lack of invariance raises a fundamental question about the quality of the resulting policy $\bar \pi$. If the "optimal" solution shifts based on different choices $\Psi$ and $\pi_{\mathrm{ref}}$, it suggests that $\bar{\pi}$ is an artifact of the objective function rather than a faithful reflection of human preferences. Consequently, in the following,  we demonstrate that the current target $\bar{\pi}$ is, in fact, technically suboptimal.

\subsection{Is $\bar{\pi}$ Optimal?}
In this subsection, we investigate the theoretical optimality of $\bar{\pi}$ in terms of maximizing $\mathcal V(\pi)$. While $\bar{\pi}$ maximizes $\mathcal V(\pi)$ over $\bar \Pi$, we show that it is generally suboptimal compared to the globally optimal policy $\pi^\ast$, defined as:
\begin{align}\label{def: an optimal policy}
    \pi^\ast \in \mathrm{argmax}_{\pi \in \Pi} \mathcal V(\pi),
\end{align}
where $\Pi = \{\pi: \mathcal X \times \mathcal Y \rightarrow \mathcal Y \}$ represents a broader policy class that conditions not only on the context $x$ but also on an auxiliary response $y'$. 
The following theorem establishes that $\pi^\ast$ is not only superior to $\bar \pi$ but also possesses the desirable invariance property.
\begin{theorem}\label{thm: optimality}
(i) $\pi^\ast$ is invariant to $\Psi$ and $\pi_{\mathrm{ref}}$; (ii)
    The following inequality holds, which implies $\pi^\ast$ is superior to $\bar \pi$.
    \begin{align*}
        &\mathcal V(\pi^\ast) = \mathbb{E}_{x \sim \rho}
\mathbb{E}_{y \sim \textcolor{red}{ \pi^\ast(\cdot\mid x, y')}, y' \sim \pi_{\mathrm{ref}}(\cdot\mid x)}
\left[
\Psi\!\big( \mathbb{P}(y \succ y' \mid x) \big)
\right]\\
&\geq \mathcal V(\bar \pi) = \mathbb{E}_{x \sim \rho}
\mathbb{E}_{y \sim \textcolor{red}{\bar \pi(\cdot\mid x)}, y' \sim \pi_{\mathrm{ref}}(\cdot\mid x)}
\left[
\Psi\!\big( \mathbb{P}(y \succ y' \mid x) \big)
\right].
    \end{align*}
    (ii) Furthermore, given a fixed $\Psi$, $\pi^\ast$ coincides with $\bar{\pi}$ if the transformed preference probability satisfies the condition that $\Psi\big( \mathbb{P}(y \succ y' \mid x) \big) \propto c(x, y) - c(x, y')$ for some function $c$.
\end{theorem}
Theorem \ref{thm: optimality} implies that the superiority of $\pi^\ast$ over $\bar{\pi}$ stems from its dependence on the comparator response $y'$, which effectively triggers a novel notion of intrinsic \textit{self-reflection} within the LLM. This capacity for self-reflection, which existing fine-tuning methods lack, is a critical property for enhancing alignment. Furthermore, this mechanism renders $\pi^\ast$ invariant to both $\Psi$ and $\pi_{\mathrm{ref}}$, ensuring that the policy targets the ground-truth human preference probability rather than optimization artifacts as $\bar \pi$. While Theorem \ref{thm: optimality}~(iii) suggests that self-reflection yields no improvement if the preference function is separable, the separable condition is restrictive as it requires a correctly specified (link) function $\Psi$. In other words, in the current framework of RLHF and DPO, $\bar \pi$ is optimal only if the BT model \eqref{eq:bt} is correctly specified. More importantly, existing literature \citep{tversky1993context, bordalo2013salience} demonstrates that human preferences are inherently non-separable and determined by the interaction between options. It is the context $x$, and the self-reflection by comparing $y$ with $y'$ that fundamentally shape the choice.
Consequently, to leverage this comparative property, we aim to learn $\pi^\ast$ from the preference data $\mathcal D_n$, and since our preference data are paired, we do not consider a larger categories of $\Pi$ beyond pairwise comparison.

\section{Intrinsic Self-reflective Preference Optimization}


In this section, we introduce our InSPO method for learning $\pi^\ast$. Thanks to the invariant property of $\pi^\ast$ as shown in Theorem \ref{thm: optimality}~(i), we consider $\Psi(q) = \log(q/(1-q))$.
First of all, we impose the following choice model.
\begin{align}\label{eqn: general reward model}
    \mathbb{P}(y_w \succ y_\ell \mid x)= \sigma(2(r(x,y_w, y_\ell) - \beta \log Z(x, y_\ell))),
\end{align}
for some generic reward function $r: \mathcal X \times \mathcal Y \times \mathcal Y \rightarrow \mathbb R$ and $Z(x,y'):=\sum_{y}\pi_{\mathrm{ref}}(y\mid x)\,
\exp\!\Big(\tfrac{1}{\beta}r(x,y,y')\Big)$.  Then by \eqref{def: an optimal policy}, 
\begin{align*}
    \pi^\ast \in &  \mathrm{argmax}_{\pi} \, 2\mathbb{E}_{x \sim \rho, y \sim \pi(\cdot \mid x, y'), y' \sim \pi_{\mathrm{ref}}(\cdot \mid x)}\left[r(x, y, y') - \beta \log Z(x, y')\right]\\
    =& \mathrm{argmax}_{\pi} \,\mathbb{E}_{x \sim \rho, y \sim \pi(\cdot \mid x, y'), y' \sim \pi_{\mathrm{ref}}(\cdot \mid x)}\left[r(x, y, y')\right],
\end{align*}
which is independent of $Z$. Therefore model assumption in \eqref{eqn: general reward model} is mild as $r$ is unspecified and can be generic. 
Then following the paradigm of RLHF, we can estimate $\pi^\ast$ via
\begin{align}\label{eqn: ppo2}
    \max_{\pi} \, \mathbb{E}_{x \sim \rho, y \sim \pi(\cdot \mid x, y'), y' \sim \pi_{\mathrm{ref}}(\cdot \mid x)}\left[r(x, y, y') \right] - \beta D_{\mathrm{KL}}(\pi \,\|\, \pi_{\mathrm{ref}}),
\end{align}
where 
\[
D_{\mathrm{KL}}(\pi \,\|\, \pi_{\mathrm{ref}})
= \mathbb{E}_{x \sim \rho, y' \sim \pi_{\mathrm{ref}}(\cdot \mid x)} \Big[ \mathrm{KL}\big(\pi(\cdot \mid y', x) \,\|\, \pi_{\mathrm{ref}}(\cdot \mid x)\big) \Big].
\]
While this is a promising approach, the reward function $r$ may be hard to estimate and PPO is known for instability. In the following, we propose a family of DPO-based approaches for learning $\pi^\ast$.


To begin with, we have the following proposition that establishes the connection between the general reward function $r$ and $\pi_r$, which is denoted as an optimal solution to \eqref{eqn: ppo2}.
\begin{theorem}\label{thm: srapo}
    Solving the optimization problem \eqref{eqn: ppo2} gives
    \begin{align}\label{eqn: DPO and PPO}
    r(x, y, y') =   & \beta\Big[
\log\tfrac{\pi_r(y\mid x,y')}{\pi_{\mathrm{ref}}(y\mid x)}
+\log Z(x,y')\Big].
    \end{align}
Furthermore, $\pi_r$ can be obtained by solving
\begin{align}\label{eqn: self-reflection DPO}
    \max_\pi\;
\mathbb{E}_{(x,y_w,y_\ell)\sim \mathcal D }
\Big[
\log \sigma \Big(
\beta\Big(
\log\tfrac{\pi(y_w\mid x,\textcolor{red}{y_\ell})}{\pi_{\mathrm{ref}}(y_w\mid x)}
-\log\tfrac{\pi(y_\ell\mid x, \textcolor{red}{y_w})}{\pi_{\mathrm{ref}}(y_\ell\mid x)}
\Big)
\Big)
\Big].
\end{align}
\end{theorem}
\noindent Then based on Theorem \ref{thm: optimality} and the preference dataset $\mathcal D_n$, InSPO estimates $\pi^\ast$ via
\begin{align}\label{eqn: empirical self-reflection DPO}
    \min_{\theta}\;
-\frac{1}{n}\sum_{i = 1}^n
\Big[
\log \sigma \Big(
\beta\Big(
\log\tfrac{\pi_\theta(y^{(i)}_w\mid x,\textcolor{red}{y^{(i)}_\ell})}{\pi_{\mathrm{ref}}(y^{(i)}_w\mid x)}
-\log\tfrac{\pi_\theta(y^{(i)}_\ell\mid x, \textcolor{red}{y^{(i)}_w})}{\pi_{\mathrm{ref}}(y^{(i)}_\ell\mid x)}
\Big)
\Big)
\Big],
\end{align}
where we parametrize the trainable policy $\pi$ by $\theta$.
Similar to DPO, the proposed InSPO based on \eqref{eqn: empirical self-reflection DPO} aligns an LLM by directly shifting probability mass toward human preferred responses and away from dispreferred responses without performing RL, without rollouts, and without training a reward model. On top of it, our method leverages the self-reflection property embedded in the preference data to further calibrate the policy towards the preferred response by contrasting with the dispreferred one. Therefore, this approach enables the learning of $\pi^\ast$. The training procedure is outlined in Algorithm~\ref{alg:src} and denote the final solution to  \eqref{eqn: empirical self-reflection DPO} as $\widehat \pi$.

During the inference, instead of first generating $y'$ from $\pi_{\mathrm{ref}}$ for self-reflection, which incurs an overhead cost, we directly deploy $\widehat \pi$ given a testing query $x$. While this introduces a distinction between the training context $(x, y')$ and the inference context $x$, we understand this under the paradigm of {Learning Using Privileged Information (LUPI)} \citep{vapnik2009new}. In the LUPI framework, the learning algorithm is provided with additional ``privileged'' information (here, the alternative response) during the training phase to stabilize the optimization landscape and accelerate the convergence. This privileged context acts as a contrastive scaffold, guiding the gradient updates for the shared weights associated with $x$. Consequently, the self-reflective capability is \textit{distilled} into the policy weights \citep{hinton2015distilling}, allowing the deployed model to retain the optimized decision boundaries even when the privileged scaffolding is removed at test time. This ensures that InSPO incurs zero inference overhead, as the self-reflective mechanism is implicitly encoded in the parameter space rather than requiring explicit rollout generation. 



\begin{algorithm}[htbp]
\caption{Self-reflection based DPO}
\label{alg:src}
\begin{algorithmic}[1]
\STATE \textbf{Input:} $\mathcal{D}_n = \{(x, y_w, y_\ell)\}_{i = 1}^n$, $\pi_{\mathrm{ref}}$, $\beta$ \qquad \textbf{Output:} $\widehat{\pi}$
\STATE \textbf{Initialize:} $\pi_\theta \leftarrow \pi_{\mathrm{ref}}$
\FOR{$(x, y_w, y_\ell) \sim \mathcal{D}_n$}
    \STATE $r_w, r_\ell \leftarrow \log \pi_{\mathrm{ref}}(y_w | x), \log \pi_{\mathrm{ref}}(y_\ell | x)$ \hfill $\triangleright$ Reference (prompt-level)
    \STATE $p_w, p_\ell \leftarrow \log \pi_{\theta}(y_w | x, y_\ell), \log \pi_{\theta}(y_\ell | x, y_w)$ \hfill $\triangleright$ Policy (cross-conditioned)
    \STATE $m \leftarrow \beta (p_w - r_w - p_\ell + r_\ell)$ \hfill $\triangleright$ Preference margin
    \STATE $\theta \leftarrow \theta - \eta \nabla_\theta [-\log \sigma(m)]$ \hfill $\triangleright$ Eq.~\eqref{eqn: empirical self-reflection DPO}
\ENDFOR
\RETURN $\widehat{\pi}$
\end{algorithmic}
\end{algorithm}


It is worth noting that our proposed formulation for $\pi^\ast$ is method-agnostic and can seamlessly integrate with standard preference optimization techniques. Table \ref{tab:obj-merged} illustrates the InSPO variants of six representative preference optimization methods, including DPO, demonstrating how each is adapted to unlock the self-reflective capability. In the next section, we study the empirical performance of each method.

\begin{table}[ht!]
\centering
\normalsize
\setlength{\tabcolsep}{4pt}
\renewcommand{\arraystretch}{1.08}
\caption{InSPO variants of six representative preference optimization methods. The Original formulations of DPO-based approaches are in black, while \q-enhanced terms are in {\color{green!40!black}green}.}
\label{tab:obj-merged}
\begin{tabular}{>{\centering\arraybackslash}m{0.12\linewidth} >{\centering\arraybackslash}m{0.84\linewidth}}
\toprule
\textbf{Method} & \textbf{Objective Function for Minimization} \\
\midrule
\textbf{DPO \citep{rafailov2023direct}} &
$\displaystyle -\log\sigma\!\Bigl(\beta\log\tfrac{\pi_\theta(y_w|x)}{\pi_{\mathrm{ref}}(y_w|x)} - \beta\log\tfrac{\pi_\theta(y_\ell|x)}{\pi_{\mathrm{ref}}(y_\ell|x)}\Bigr)$ \\ 
\cmidrule(lr){2-2}
&
$\displaystyle -\log\sigma\!\Bigl(\beta\log\tfrac{{\color{green!40!black}\pi_\theta(y_w|y_\ell,x)}}{\pi_{\mathrm{ref}}(y_w|x)} - \beta\log\tfrac{{\color{green!40!black}\pi_\theta(y_\ell|y_w,x)}}{\pi_{\mathrm{ref}}(y_\ell|x)}\Bigr)$ \\
\midrule
\textbf{IPO~\citep{azar2024general}} &
$\displaystyle \Bigl(\log\tfrac{\pi_\theta(y_w|x)}{\pi_{\mathrm{ref}}(y_w|x)} - \log\tfrac{\pi_\theta(y_\ell|x)}{\pi_{\mathrm{ref}}(y_\ell|x)} - \tfrac{1}{2\tau}\Bigr)^{\!2}$ \\
\cmidrule(lr){2-2}
&
$\displaystyle \Bigl(\log\tfrac{{\color{green!40!black}\pi_\theta(y_w|y_\ell,x)}}{\pi_{\mathrm{ref}}(y_w|x)} - \log\tfrac{{\color{green!40!black}\pi_\theta(y_\ell|y_w,x)}}{\pi_{\mathrm{ref}}(y_\ell|x)} - \tfrac{1}{2\tau}\Bigr)^{\!2}$ \\
\midrule
\textbf{RDPO~\citep{park2024disentangling}} &
$\displaystyle -\log\sigma\!\Bigl(\beta[\log\tfrac{\pi_\theta(y_w|x)}{\pi_{\mathrm{ref}}(y_w|x)} - \log\tfrac{\pi_\theta(y_\ell|x)}{\pi_{\mathrm{ref}}(y_\ell|x)}] + \alpha(|y_w|-|y_\ell|)\Bigr)$ \\
\cmidrule(lr){2-2}
&
$\displaystyle -\log\sigma\!\Bigl(\beta[\log\tfrac{{\color{green!40!black}\pi_\theta(y_w|y_\ell,x)}}{\pi_{\mathrm{ref}}(y_w|x)} - \log\tfrac{{\color{green!40!black}\pi_\theta(y_\ell|y_w,x)}}{\pi_{\mathrm{ref}}(y_\ell|x)}] + \alpha(|y_w|-|y_\ell|)\Bigr)$ \\
\midrule
\textbf{ORPO~\citep{hong2024orpo}} &
$\displaystyle -\log p_\theta(y_w\mid x)\;-\;\lambda\,\log\sigma\!\Bigl(\logit\!\bigl (p_\theta(y_w\mid x)\bigr)\;-\;\logit\!\bigl (p_\theta(y_\ell\mid x)\bigr)\Bigr)$ \\
\cmidrule(lr){2-2}
&
$\displaystyle -\log {\color{green!40!black}p_\theta(y_w\mid y_\ell,x)}\;-\;\lambda\,\log\sigma\!\Bigl(\logit\!\bigl({\color{green!40!black}p_\theta(y_w\mid y_\ell,x)}\bigr)\;-\;\logit\!\bigl({\color{green!40!black}p_\theta(y_\ell\mid y_w,x)}\bigr)\Bigr)$ \\
\midrule
\textbf{SimPO~\citep{meng2024simpo}} &
$\displaystyle -\log\sigma\!\Bigl(\beta\tfrac{1}{|y_w|}\log\pi_\theta(y_w|x) - \beta\tfrac{1}{|y_\ell|}\log\pi_\theta(y_\ell|x) - \gamma\Bigr)$ \\
\cmidrule(lr){2-2}
&
$\displaystyle -\log\sigma\!\Bigl(\beta\tfrac{1}{|y_w|}\log{\color{green!40!black}\pi_\theta(y_w|y_\ell,x)} - \beta\tfrac{1}{|y_\ell|}\log{\color{green!40!black}\pi_\theta(y_\ell|y_w,x)} - \gamma\Bigr)$ \\
\bottomrule
\end{tabular}
\vspace{0.3em}
\footnotesize
$p_\theta(y|{\cdot}){=}\exp(\tfrac{1}{|y|}\log\pi_\theta(y|{\cdot}))$;
$\beta,\tau,\alpha,\lambda,\gamma$: hyperparameters.
\end{table}

\begin{figure*}[!t]
    \centering
    \begin{minipage}[t]{.33\textwidth}\centering
        \includegraphics[width=\linewidth]{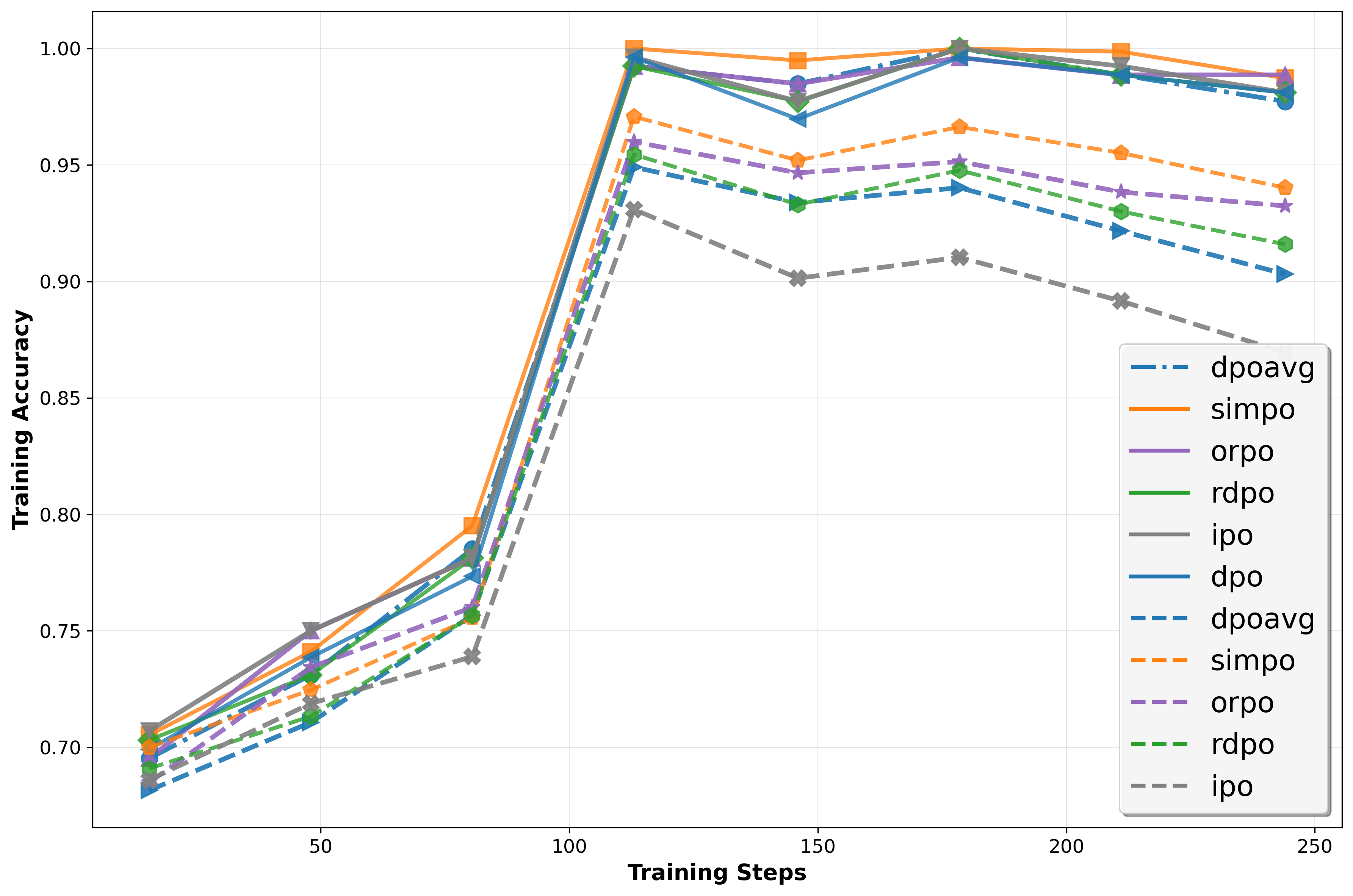}
        \vspace{2pt}\footnotesize (a) Preference Accuracy
    \end{minipage}\hfill
    \begin{minipage}[t]{.33\textwidth}\centering
        \includegraphics[width=\linewidth]{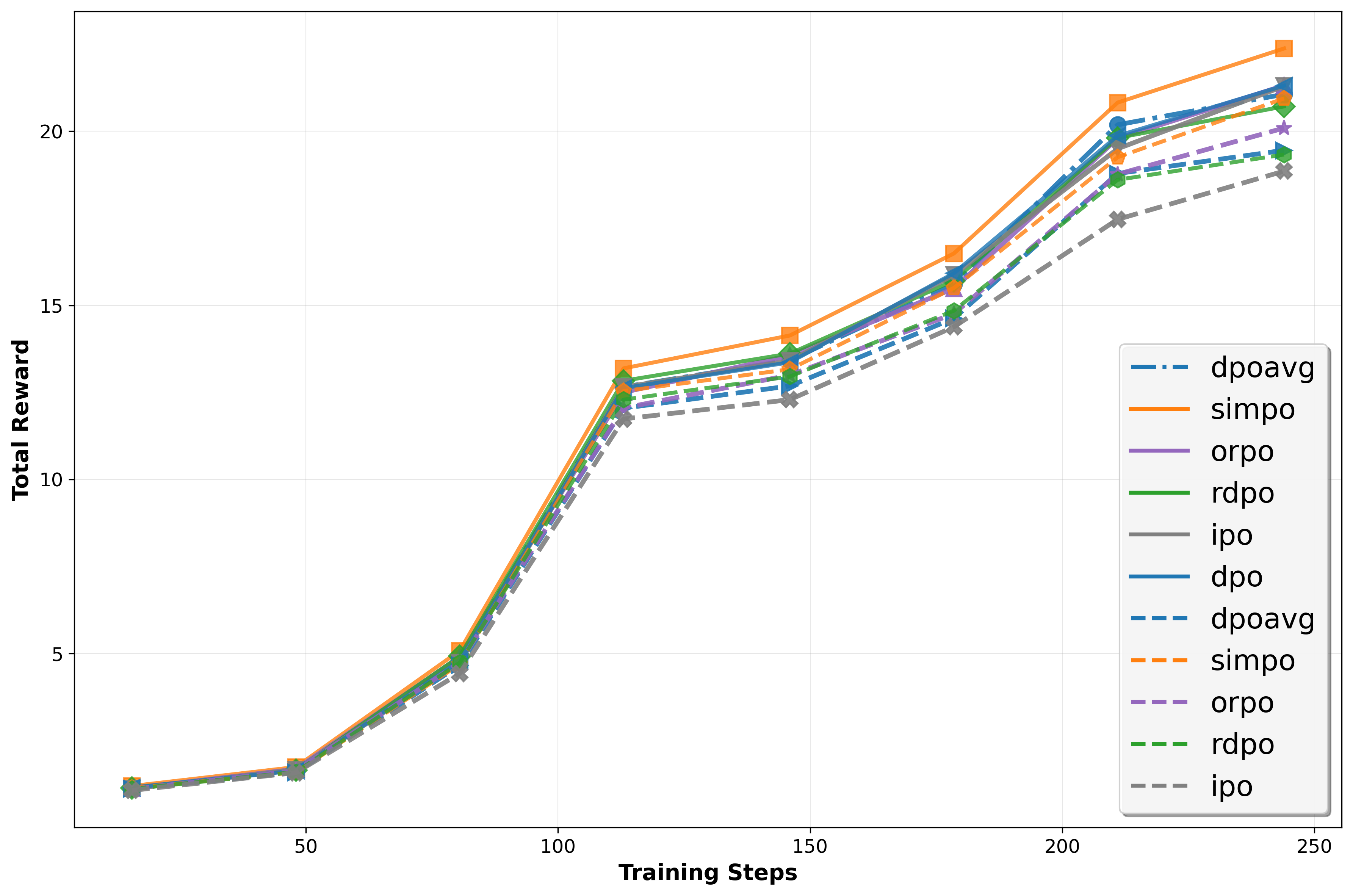}
        \vspace{2pt}\footnotesize (b) Reward 
    \end{minipage}\hfill
    \begin{minipage}[t]{.33\textwidth}\centering
        \includegraphics[width=\linewidth]{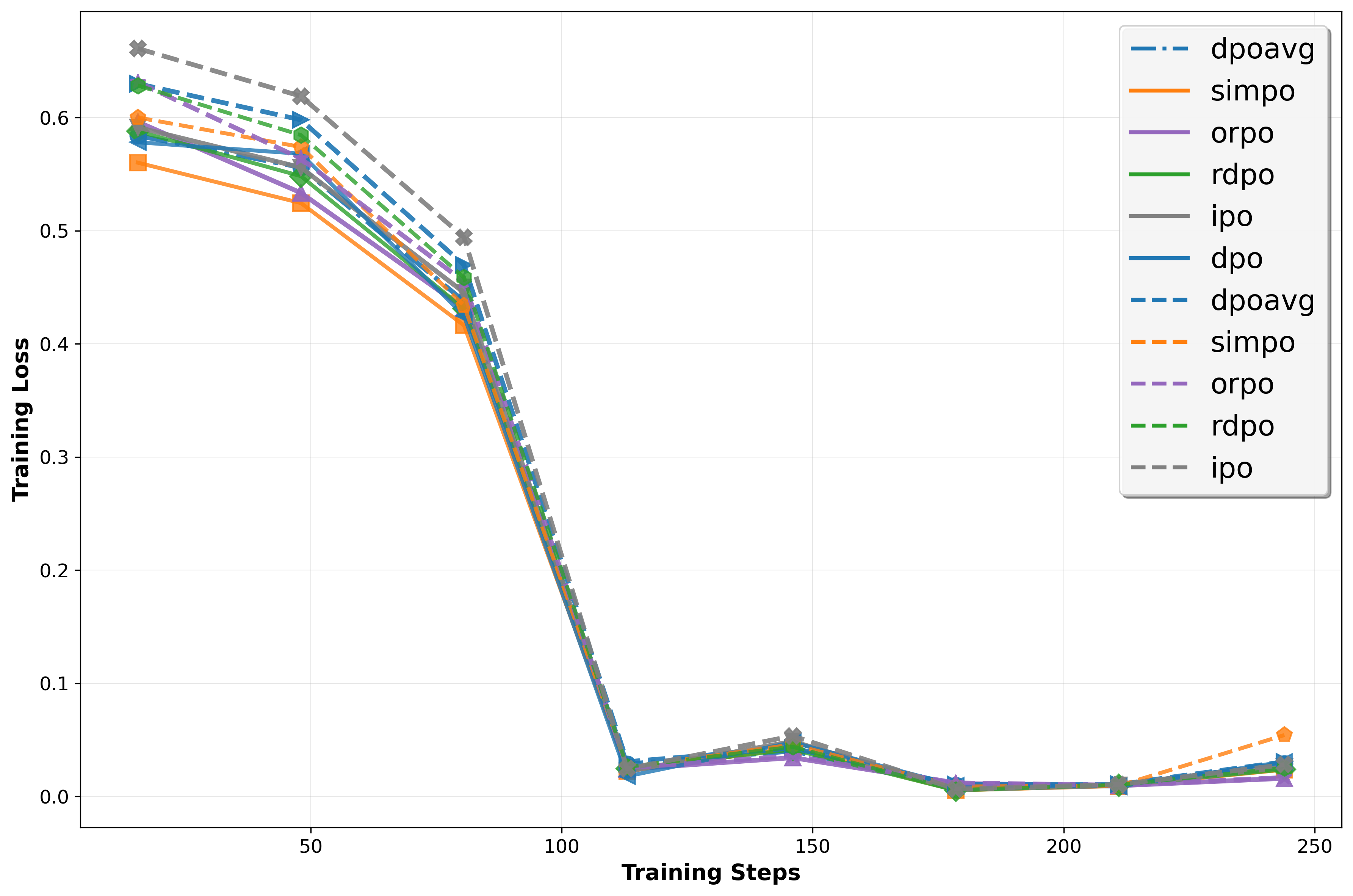}
        \vspace{2pt}\footnotesize (c) Training Loss
    \end{minipage}
    \caption{Training dynamics of \q~methods (solid) versus baselines (dashed). Our \q~exhibits stable optimization with smooth loss convergence, consistent accuracy improvement, and enhanced reward margins without requiring additional on-policy rollouts.}
    \label{fig:training-curves}
\end{figure*}

\begin{table*}[!t]
\centering
\caption{Benchmark performance on AlpacaEval2, Arena-Hard, and MT-Bench for Mistral-Instruct (7B) and Llama-3-Instruct (8B). \colorbox{blue!8}{Shaded rows} denote continuation-conditioned variants with deltas computed relative to their corresponding baselines. Bold values indicate the highest performance in each metric. All methods employ DPO-family objectives with identical training configurations.}
\label{tab:main}
\setlength{\tabcolsep}{2.5pt}
\small
\begin{tabular}{l cc c cc cc c cc}
\toprule
& \multicolumn{5}{c}{\textbf{Mistral-Instruct (7B)}} & \multicolumn{5}{c}{\textbf{Llama-3-Instruct (8B)}} \\
\cmidrule(lr){2-6}\cmidrule(lr){7-11}
\textbf{Method} & \multicolumn{2}{c}{AlpacaEval2} & Arena-Hard & \multicolumn{2}{c}{MT-Bench}
                & \multicolumn{2}{c}{AlpacaEval2} & Arena-Hard & \multicolumn{2}{c}{MT-Bench} \\
\cmidrule(lr){2-3}\cmidrule(lr){4-4}\cmidrule(lr){5-6}\cmidrule(lr){7-8}\cmidrule(lr){9-9}\cmidrule(lr){10-11}
& LC (\%) & WR (\%) & WR (\%) & GPT-4o & GPT-4 & LC (\%) & WR (\%) & WR (\%) & GPT-4o & GPT-4 \\
\midrule\midrule
Baseline     & 17.1 & 14.7 & 12.6 & 6.2 & 7.5   & 26.0 & 25.3 & 22.3 & 6.9 & 8.1 \\
\midrule\midrule
\multicolumn{11}{l}{\textit{Reward-Free Preference Optimization}} \\
RRHF      & 25.3 & 24.8 & 18.1 & 6.5 & 7.6   & 31.3 & 28.4 & 26.5 & 6.7 & 7.9 \\
SLiC-HF  & 24.1 & 24.6 & 18.9 & 6.5 & 7.8   & 26.9 & 27.5 & 26.2 & 6.8 & 8.1 \\
CPO        & 23.8 & 28.8 & 22.6 & 6.3 & 7.5   & 28.9 & 32.2 & 28.8 & 7.0 & 8.0 \\
KTO     & 24.5 & 23.6 & 17.9 & 6.4 & 7.7   & 33.1 & 31.8 & 26.4 & 6.9 & 8.2 \\
\midrule\midrule
\multicolumn{11}{l}{\textit{Preference Optimization with Self-Reflection Conditioning}} \\
IPO & 20.3 & 20.3 & 16.2 & 6.4 & 7.8   
                    & 35.6 & 35.6 & 30.5 & 7.0 & 8.3 \\
\oursrow \quad\textbf{+ \m}   & 28.8 \inc{8.5} & 27.9 \inc{7.6} & 18.5 \inc{2.3} & 6.5 \inc{0.1} & 7.9 \inc{0.1}
                    & 37.7 \inc{2.1} & 41.3 \inc{5.7} & 32.8 \inc{2.3} & 7.1 \inc{0.1} & 8.5 \inc{0.2} \\
\addlinespace
ORPO & 24.5 & 24.9 & 20.8 & 6.4 & 7.7   
                    & 28.5 & 27.4 & 25.8 & 6.8 & 8.0 \\
\oursrow \quad\textbf{+ \m}   & 24.6 \inc{0.1} & 24.7 \dec{0.2} & 21.5 \inc{0.7} & 6.3 \dec{0.1} & 7.8 \inc{0.1}
                    & 35.7 \inc{7.2} & 40.3 \inc{12.9} & 25.5 \dec{0.3} & 6.8 \sta{0.0} & 8.0 \sta{0.0} \\
\addlinespace
R-DPO & 27.3 & 24.5 & 16.1 & 6.2 & 7.5   
                    & 41.1 & 37.8 & 33.1 & 7.0 & 8.0 \\
\oursrow \quad\textbf{+ \m}   & 29.1 \inc{1.8} & 27.6 \inc{3.1} & 19.0 \inc{2.9} & 6.4 \inc{0.2} & 7.6 \inc{0.1}
                    & 41.0 \dec{0.1} & 43.8 \inc{6.0} & 35.0 \inc{1.9} & 7.1 \inc{0.1} & 8.1 \inc{0.1} \\
\addlinespace
DPO & 26.8 & 24.9 & 16.3 & 6.3 & 7.6   
                    & 40.3 & 37.9 & 32.6 & 7.0 & 8.0 \\
\oursrow \quad\textbf{+ \m}   & 29.9 \inc{3.1} & 29.4 \inc{4.5} & 23.9 \inc{7.6} & 6.5 \inc{0.2} & 7.8 \inc{0.2}
                    & 40.3 \sta{0.0} & 41.1 \inc{3.2} & 35.3 \inc{2.7} & 7.1 \inc{0.1} & 8.1 \inc{0.1} \\
\addlinespace
SimPO  & 32.1 & \textbf{34.6} & 21.0 & 6.6 & 7.6   
                    & \textbf{44.5} & 40.5 & 33.8 & 7.0 & 8.0 \\
\oursrow \quad\textbf{+ \m}   & \textbf{32.7} \inc{0.6} & 34.5 \dec{0.1} & \textbf{24.4} \inc{3.4} & \textbf{6.6} \sta{0.0} & \textbf{7.7} \inc{0.1}
                    & 44.5 \sta{0.0} & \textbf{46.6} \inc{6.1} & \textbf{35.9} \inc{2.1} & \textbf{7.1} \inc{0.1} & \textbf{8.2} \inc{0.2} \\
\bottomrule
\end{tabular}
\vspace{0.25em}

\footnotesize
LC: length-controlled win rate; WR: win rate; AE: AlpacaEval2; AH: Arena-Hard.
All methods initialize from the same instructed checkpoint within each model family. The proposed
\textbf{\m} enables self-reflection via symmetric cross-conditioning on preference pairs.
\end{table*}

\section{Experiments}
\subsection{Experimental Setup}
\textbf{Models and data.} We employ Mistral-7B-Instruct-v0.2 and Llama-3-8B-Instruct as base models, initializing all methods from identical checkpoints within each family. Training data is sourced from UltraFeedback~\citep{cui2023ultrafeedback}, containing approximately 60K preference pairs $(x; y_w, y_\ell)$ with $y_w \succ y_\ell$ after deduplication and safety filtering. The data generating process follows exactly from SimPO~\citep{meng2024simpo}.

\textbf{Benchmarks.} We evaluate on three widely-adopted benchmarks. 
\textbf{AlpacaEval 2}~\citep{dubois2024length} contains 805 diverse 
instructions; we report both standard win rates (WR) and length-controlled 
win rates (LC) to mitigate verbosity bias. 
\textbf{Arena-Hard}~\citep{li2024crowdsourced} features 500 challenging 
queries that test advanced reasoning; we report WR alongside its 
style-controlled variant (SC) to account for stylistic preferences. 
\textbf{MT-Bench}~\citep{zheng2023judging} comprises 80 multi-turn questions spanning eight capability categories, scored on a 10-point scale.

\textbf{Baselines.} We compare \q-enhanced variants named \m-DPO, \m-SimPO, \m-IPO, \m-RDPO, \m-ORPO against their standard counterparts and additional baselines: RRHF~\citep{yuan2023rrhf}, SLiC-HF~\citep{zhao2023slic}, CPO~\citep{xu2024contrastive}, and KTO~\citep{ethayarajh2024kto}. For fair comparison, baseline results are obtained by evaluating publicly released checkpoints from SimPO~\citep{meng2024simpo} on identical benchmark versions. All \q-enhanced methods are implemented using the OpenRLHF framework~\citep{hu2024openrlhf} and trained for three epochs with AdamW optimizer, learning rate $5\times10^{-7}$, cosine schedule with 10\% warmup, and maximum context length 4096. Method-specific hyperparameters follow exact configurations from prior works.

\textbf{Inference.} All models use a standard autoregressive generation $\hat{y} \sim \widehat \pi(\cdot\mid x)$ with nucleus sampling at $p=0.95$ and temperature $0.7$, introducing no additional computational overhead versus baseline methods.

\subsection{Main Results}
Table~\ref{tab:main} reports comprehensive benchmark results. Controlled metrics show substantial improvements of InSPO. On AlpacaEval(AE) LC, \q~yields 0.6--8.5 point gains across the DPO family, with the largest improvements observed on \m-IPO: +8.5 points for Mistral-Instruct and +2.1 points for Llama-3-Instruct. Arena-Hard(AH) results follow a similar pattern, with improvements ranging from 0.7 to 7.6 points. Notably, \m-DPO achieves a 7.6-point gain on Mistral-Instruct and 2.7 points on Llama-3-Instruct, demonstrating consistent benefits of enabling intrinsic self-reflection through alternative conditioning.
The training dynamic curves in Figure~\ref{fig:training-curves} further enhance the persuasiveness of the results.

The gains extend beyond controlled metrics: \m-DPO and \m-SimPO improve raw win rates by 3--6 points on AlpacaEval 2 and 2--3 points on Arena-Hard, indicating genuine quality improvements rather than mere verbosity reduction. MT-Bench results corroborate this finding, with \m-enhanced methods achieving 0.1--0.2 point gains across different judge models, demonstrating improvements where alternative response context provides valuable learning signals.

The relative ranking among DPO family members remains stable under \q~enhancement—SimPO variants consistently lead, followed by DPO and R-DPO, with IPO and ORPO showing more variable performance—suggesting that \q~amplifies rather than disrupts the inherent strengths of each base method. Compared to alternative baselines beyond the DPO family such as RRHF, SLiC-HF, CPO, and KTO, our best \m-enhanced variant \m-SimPO establishes competitive results on controlled metrics while maintaining comparable raw performance.

\subsection{Analysis and Insights}

\begin{table}[t]
\centering
\small
\setlength{\tabcolsep}{5pt}
\renewcommand{\arraystretch}{1.15}
\caption{Ablation study on candidate length constraints and context windows using Llama-3-8B. MaxLen limits the total sequence length of $(x, y_\ell, y_w)$; Draft Cap $\alpha$ enforces $|y_\ell| \le \alpha \cdot \text{MaxLen}$ for the dispreferred candidate only.}
\label{tab:length-ablation}
\begin{tabular}{cl cccc}
\toprule
\textbf{MaxLen} & \textbf{Draft Cap} & \textbf{AE-LC} & \textbf{AE-WR} & \textbf{AH-SC} & \textbf{AH-WR} \\
\midrule
\multirow{4}{*}{1024}
& 30\% & 39.1 & 37.4 & 30.8 & 32.1 \\
& 40\% & 39.8 & \textbf{38.2} & \textbf{31.2} & 32.8 \\
& 50\% & \textbf{39.9} & 37.9 & 30.5 & \textbf{33.1} \\
& None & 38.2 & 36.1 & 29.8 & 31.2 \\
\midrule
\multirow{4}{*}{2048}
& 30\% & 39.9 & 38.6 & 32.3 & 34.1 \\
& 40\% & \textbf{40.3} & \textbf{39.2} & 32.9 & \textbf{34.7} \\
& 50\% & 40.1 & 38.9 & \textbf{33.1} & 34.3 \\
& None & 39.1 & 37.8 & 31.5 & 33.2 \\
\midrule
\multirow{4}{*}{4096}
& 30\% & 40.1 & 39.5 & 33.2 & 35.4 \\
& 40\% & \textbf{40.5} & \textbf{39.9} & \textbf{33.8} & \textbf{35.9} \\
& 50\% & 40.2 & 39.7 & 33.4 & 35.6 \\
& None & 39.8 & 38.9 & 32.8 & 34.9 \\
\bottomrule
\end{tabular}
\end{table}

\begin{table}[t]
\centering
\small
\setlength{\tabcolsep}{4pt}
\renewcommand{\arraystretch}{1.15}
\caption{Comparison of conditioning strategies and simple averaging for length control. All experiments use MaxLen=4096 with 40\% draft cap. Deltas are computed relative to the no-conditioning baseline.}
\label{tab:conditioning-comparison}
\begin{tabular}{ll ccc}
\toprule
\textbf{Model} & \textbf{Strategy} & \multicolumn{2}{c}{\textbf{AlpacaEval 2}} & \textbf{Arena-Hard} \\
\cmidrule(lr){3-4}\cmidrule(lr){5-5}
& & LC (\%) & WR (\%) & WR (\%) \\
\midrule
Mistral-7B & DPO (baseline) & 26.8 & 24.9 & 16.3 \\
& One-sided & 28.7 \inc{1.9} & 27.1 \inc{2.2} & 19.2 \inc{2.9} \\
& Symmetric & 29.9 \inc{3.1} & 29.4 \inc{4.5} & 23.9 \inc{7.6} \\
& Averaged & 29.6 \inc{2.8} & 29.7 \inc{4.8} & 25.7 \inc{9.4} \\
\midrule
Llama-3-8B & DPO (baseline) & 40.3 & 37.9 & 32.6 \\
& One-sided & 43.4 \inc{3.1} & 40.2 \inc{2.3} & 34.1 \inc{1.5} \\
& Symmetric & 40.3 \inc{0.0} & 41.1 \inc{3.2} & 35.3 \inc{2.7} \\
& Averaged & 44.1 \inc{3.8} & 41.5 \inc{3.6} & 35.5 \inc{2.9} \\
\bottomrule
\end{tabular}
\end{table}

We conduct systematic ablation studies to understand the mechanisms underlying sequence-level conditioning and explore its interactions with key design choices. 
Unless otherwise specified, all experiments use \m-DPO with the Llama-3-8B-Instruct model.

\textbf{Impact of candidate length and context window.} 
We systematically investigate this by varying both the maximum context window and the candidate length cap $\alpha$. Specifically, \textit{MaxLen} constrains the total sequence length of the entire training tuple 
$(x, y_\ell, y_w)$, while the \textit{Draft Cap} $\alpha$ enforces $|y_\ell| \le \alpha \cdot \text{MaxLen}$ for the dispreferred candidate alone. Crucially, larger MaxLen values not only permit longer $y_\ell$, 
but also provide more headroom for the instruction $x$ and preferred response $y_w$ to remain untruncated.

Table~\ref{tab:length-ablation} presents results across three context window sizes (1024, 2048, 4096) and four candidate length constraints (30\%, 40\%, 50\%, uncapped).
Three key findings emerge. 
\textit{First}, longer context windows consistently improve performance, with 4096-token contexts yielding the best results across all metrics (40.3\% LC, 39.9\% WR on AlpacaEval 2). This suggests that providing sufficient space for the complete $(x, y_\ell, y_w)$ sequence without truncation is crucial for effective sequence-level learning.
\textit{Second}, moderate draft length caps (40\% of MaxLen) achieve optimal performance, balancing context richness with training stability. Uncapped drafts slightly underperform, likely due to occasional extremely long dispreferred responses that dominate the context and destabilize gradient flow. 
\textit{Third}, the benefits of longer contexts are most pronounced on challenging benchmarks: on Arena-Hard, the 4096-token setting outperforms 1024 by 2.6 WR points, indicating that complex reasoning tasks particularly benefit from richer candidate context.

\textbf{Conditioning strategies and length control.} 
Our main results in Table~\ref{tab:main} show that sequence-level conditioning consistently improves raw win rates, and yields measurable improvements under length-controlled evaluation.
This observation, combined with the finding that longer contexts enable better performance, motivates us to explore whether different conditioning strategies can better balance quality improvements with length control.
We compare four variants:
\begin{description}[nosep, leftmargin=1.5em, style=sameline]
    \item[(i)] \textit{No conditioning} (standard DPO baseline).
    \item[(ii)] \textit{One-sided conditioning}: $\mathcal{L}_{\text{one-sided}} = -\log\sigma\big(\beta[\log\pi_\theta(y_w|x,y_\ell)/\pi_{\mathrm{ref}}(y_w |x) - \log\pi_\theta(y_\ell|x,y_\ell)/\pi_{\mathrm{ref}}(y_\ell |x)]\big)$.
    \item[(iii)] \textit{Cross-conditioning (\q)}: $\mathcal{L}_{\text{sym}} = -\log\sigma\big(\beta[\log\pi_\theta(y_w|x,y_\ell)/\pi_{\mathrm{ref}}(y_w |x) - \log\pi_\theta(y_\ell|x,y_w)/\pi_{\mathrm{ref}}(y_\ell |x)]\big)$.
    \item[(iv)] \textit{Empirical averaged}: $\mathcal{L}_{\text{avg}} = \frac{1}{2}\mathcal{L}_{\text{DPO}} + \frac{1}{2}\mathcal{L}_{\text{sym}}$ to mitigate length bias.
\end{description}

Table~\ref{tab:conditioning-comparison} presents comprehensive results across two model families.
Symmetric cross-conditioning achieves strong performance, substantially outperforming the baseline on Arena-Hard by +7.6 WR points for Mistral-7B and +2.7 points for Llama-3-8B.
Notably, even one-sided conditioning provides meaningful improvements (+3.1 LC points on AlpacaEval 2 for Llama-3-8B), confirming that sequence-level information provides value regardless of symmetry.

The simple averaging strategy offers a practical mechanism for length control without requiring hyperparameter tuning.
By equally weighting pairwise and sequence-level objectives, the averaged approach achieves favorable trade-offs: the best Arena-Hard performance on Mistral-7B (+9.4 WR points) while maintaining competitive LC metrics, and +3.8 LC points with +3.6 WR points on AlpacaEval 2 for Llama-3-8B.
This demonstrates that practitioners can balance competing objectives through straightforward combination.
Overall, full symmetric conditioning remains the preferred choice when optimizing for raw quality, while averaging provides a simple yet effective alternative when length considerations are important.

\textbf{Computational overhead.} 
We quantify the computational cost introduced by sequence-level conditioning across Mistral-7B, Llama-3-8B, and Gemma2-9B.
Table~\ref{tab:practical-considerations} summarizes the key characteristics.
Sequence-level conditioning incurs 18--25\% training time overhead due to processing longer inputs $(x \oplus y_\ell)$ and $(x \oplus y_w)$, with detailed breakdowns provided in Appendix~\ref{sec:further}.
Critically, \textit{inference remains identical to standard DPO with zero latency overhead}, as models generate via standard sampling $\hat{y} \sim \pi_\theta(\cdot \mid x)$ without requiring explicit draft-revise loops.

\begin{table}[H]
\centering
\small
\setlength{\tabcolsep}{6pt}
\renewcommand{\arraystretch}{1.1}
\caption{Computational comparison between DPO and \m-DPO}
\label{tab:practical-considerations}
\begin{tabular}{lcc}
\toprule
& \textbf{DPO} & \textbf{\m-DPO} \\
\midrule
Training overhead & +0\% & +21\% \\
Max sequence length & 2048 & 4096 \\
Inference overhead & +0\% & +0\% \\
Draft-revise at inference & No & No \\
\bottomrule
\end{tabular}
\end{table}

\section{Related Work}

\subsection{Reinforcement Learning from Human Feedback}
RLHF is a fundamental technique for aligning large language models with human preferences and values~\citep{bai2022training}. 
The classical RLHF pipeline typically comprises three phases: supervised fine-tuning, reward model training, and policy optimization~\citep{dong2024rlhf,chen2024deep,tang2023edge,yu2025optimizing,chang2024mixed}. 
Proximal Policy Optimization \citep{schulman2017proximal} is widely used in the policy optimization stage, where the policy model is trained to maximize the learned reward while staying close to the reference model through KL regularization. 
The RLHF framework has been successfully applied to various applications, including mitigating toxicity, ensuring safety, enhancing helpfulness, and improving model reasoning abilities~\citep{chen2025tunable,chen2025neurosymbolic,li2025dual,li2025sfmdiffusion}. 
However, recent work has highlighted challenges across the PPO-based RLHF pipeline, from preference pattern to model training~\citep{ahmadian2024back,xu2024dpo}. 
Additionally, RLHF can lead to biased outcomes, such as verbose outputs and length exploitation. Given the complexity and optimization challenges of online preference optimization algorithms, researchers have been exploring more efficient and simpler alternative offline algorithms~\cite{rafailov2023direct,zeng2024token,zhang2025network}.

\subsection{Direct Preference Optimization and Variants}
Direct Preference Optimization is a notable offline preference optimization approach that eliminates the need for explicit reward modeling by reparameterizing the reward function, achieving remarkable simplicity and training stability compared to PPO-based methods~\citep{rafailov2023direct,ivison2024unpacking}. 
DPO directly learns a policy model from preference data by expressing preferences through policy and reference model ratios, avoiding the online optimization of classical RLHF. 
Building on DPO, numerous variants have been proposed to address specific limitations or enhance performance. 
Identity Preference Optimization (IPO) introduces a theoretically grounded framework that avoids DPO's assumption of replacing pairwise preferences with pointwise rewards, using a squared loss objective instead~\citep{azar2024general}. 
Contrastive Preference Optimization (CPO) incorporates sequence likelihood as reward and trains alongside a supervised fine-tuning objective for improved calibration~\citep{xu2024contrastive}. 
Kahneman-Tversky Optimization (KTO) extends preference optimization to non-paired data by modeling preferences through prospect-theoretic optimization~\citep{ethayarajh2024kto}.
Regularized DPO (R-DPO) adds explicit length regularization to prevent length exploitation~\citep{park2024disentangling}.
Simple Preference Optimization (SimPO) proposes a reference-free reward formulation using length-normalized average log probability with a target reward margin, eliminating the need for a reference model during training~\citep{meng2024simpo}.

\section{Conclusion}

In this work, we have revealed and addressed two critical limitations in current preference optimization frameworks: the lack of theoretical invariance to modeling choices and the suboptimality arising from treating response generation in isolation. We showed that standard methods like RLHF and DPO produce aligned policies that are often artifacts of parameterization rather than true reflections of human intent, while failing to leverage the comparative nature of preference data for optimal performance. To overcome these challenges, we introduced novel InSPO, a generic framework that unlocks the model's latent capacity for self-reflection. By deriving a globally optimal policy that conditions on both the context and the alternative response, InSPO achieves a target that is superior to existing baselines and guaranteed to be invariant to the choice of scalarization function or reference policy. Practically, our method serves as a plug-and-play enhancement that induces dense reward shaping during training, guiding the model to avoid specific failure modes found in rejected responses. Crucially, InSPO distills this comparative reasoning into the policy itself, delivering these gains with zero extra inference overhead. Our comprehensive experiments on a range of DPO-family baselines confirm that InSPO consistently improves alignment performance across diverse benchmarks and scales effectively with model size. By decoupling the alignment goal from modeling constraints and unleashing intrinsic self-reflection, InSPO offers a more robust, theoretically grounded, and computationally efficient path toward human-aligned AI. Future work may explore extending this self-reflective paradigm to online iterative training or conversational settings.

\bibliographystyle{plainnat}
\bibliography{aaai2026}

\appendix
\newpage
\onecolumn

\section*{Appendix}

\appendix

\section{Notation}
\label{app:notation}
\begin{table}[ht!]
\centering
\normalsize
\setlength{\tabcolsep}{6pt}
\renewcommand{\arraystretch}{1.25}
\begin{tabular}{ll|ll}
\toprule
\textbf{Symbol} & \textbf{Meaning} & \textbf{Symbol} & \textbf{Meaning} \\
\midrule
\multicolumn{4}{l}{\textit{Basic Elements}} \\
$x$ & Prompt / input context & $y$ & Response sequence \\
$y_w, y_\ell$ & Preferred and dispreferred & $y^+, y^-$ & Winner and loser (Appendix) \\
$|y|$ & Response length (tokens) & $y_t$ & Token at position $t$ \\
\midrule
\multicolumn{4}{l}{\textit{Models and Policies}} \\
$\pi_\theta(\cdot\mid x)$ & Trainable policy & $\pi_\theta(y\mid x, y_\ell)$ & Sequence-conditioned policy \\
$\pi_{\mathrm{ref}}(\cdot\mid x)$ & Reference policy & $p_\theta(y\mid c)$ & Length-normalized probability \\
\midrule
\multicolumn{4}{l}{\textit{Reward and Scoring}} \\
$r(x,y,y')$ & Reward function & $m$ & Preference margin \\
$\sigma(z)$ & Sigmoid $1/(1+e^{-z})$ & & \\
\midrule
\multicolumn{4}{l}{\textit{Hyperparameters}} \\
$\beta>0$ & Inverse temperature & $\gamma\ge 0$ & Target margin (SimPO) \\
$\alpha$ & Length coeff. (R-DPO) & $\lambda$ & Odds ratio weight (ORPO) \\
$\tau$ & Temperature (IPO) & & \\
\midrule
\multicolumn{4}{l}{\textit{Datasets and Loss Functions}} \\
$\mathcal{D}_n$ & Preference dataset & $\mathcal{X}, \mathcal{Y}$ & Prompt and response spaces \\
$n$ & Dataset size & $(x^{(i)}, y_w^{(i)}, y_\ell^{(i)})$ & $i$-th training example \\
$\mathcal{L}_{\mathrm{DPO}}$ & DPO loss & $\mathcal{L}_{\mathrm{bi}}$ & Bidirectional loss \\
\midrule
\multicolumn{4}{l}{\textit{Technical Terms}} \\
$Z(x,y')$ & Partition function & $\mathbb{P}(y_w \succ y_\ell \mid x)$ & Preference probability \\
$\mathbb{E}_{(x,y_w,y_\ell)\sim\mathcal{D}}[\cdot]$ & Expectation over data distribution & $D_{\mathrm{KL}}$ & KL divergence  \\
$\rho$ & Context distribution & \\
\midrule
\multicolumn{4}{l}{\textit{Evaluation Metrics and Operators}} \\
LC & Length-controlled WR & WR & Raw win rate \\
SC & Style-controlled WR & $\succ$ & Preference relation \\
$\oplus$ & Concatenation & & \\
\bottomrule
\end{tabular}
\caption{Comprehensive notation and symbols used throughout the paper.}
\label{tab:notation}
\end{table}

\section{Technical Proofs}
\label{app:derivations}

\textbf{Proof of Proposition \ref{prop: invariant}}
\begin{proof}
    Recall that the objective function is given by:
    \[
    \mathcal V(\pi) = \mathbb E_{x\sim\rho} \left\{ \mathbb E_{y\sim\pi(\cdot|x)} \left[ \sum_{y'} \pi_{\mathrm{ref}}(y'|x) \, \Psi\big( \mathbb P(y \succ y'|x) \big) \right] \right\}.
    \]
    For a fixed input $x$, maximizing $\mathcal V(\pi)$ is equivalent to choosing the response $y$ that maximizes the inner expectation. Define the score of a candidate response $y$ as:
    \[
    \mathcal S(y; \Psi, \pi_{\mathrm{ref}}) \triangleq \sum_{y'} \pi_{\mathrm{ref}}(y'|x) \, \Psi\big( \mathbb P(y \succ y'|x) \big).
    \]
    Then, by the definition of $\bar{\pi}$, we have:
    \[
    \bar{\pi}(y|x) = \begin{cases} 
    1, & \text{if } y = y_{\max}, \text{ where } y_{\max} = \mathrm{argmax}_{y} \mathcal S(y; \Psi, \pi_{\mathrm{ref}}) \\
    0, & \text{otherwise,}
    \end{cases}
    \]
    assuming the uniqueness of the maximization.
    
    To prove that $\bar{\pi}$ depends on $\Psi$ and $\pi_{\mathrm{ref}}$, we construct two counter-examples using two candidate responses $\bar{y}_1, \bar{y}_2$ and a reference support set $\{y'_a, y'_b\}$. Let the win-rates be:
    \begin{itemize}
        \item Candidate $\bar{y}_1$: $\mathbb P(\bar{y}_1 \succ y'_a) = 0.9$ and $\mathbb P(\bar{y}_1 \succ y'_b) = 0.2$.
        \item Candidate $\bar{y}_2$: $\mathbb P(\bar{y}_2 \succ y'_a) = 0.56$ and $\mathbb P(\bar{y}_2 \succ y'_b) = 0.56$.
    \end{itemize}

    \textbf{1. Dependence on $\Psi$ (fixing $\pi_{\mathrm{ref}}$):} Let $\pi_{\mathrm{ref}}$ be uniform: $\pi_{\mathrm{ref}}(y'_a|x) = \pi_{\mathrm{ref}}(y'_b|x) = 0.5$.
    \begin{itemize}
        \item \textit{Using Identity mapping} $\Psi_{\mathrm{id}}(q) = q$:
        \begin{align*}
            \mathcal S(\bar{y}_1; \Psi_{\mathrm{id}}, \pi_{\mathrm{ref}}) &= 0.5(0.9) + 0.5(0.2) = 0.55 \\
            \mathcal S(\bar{y}_2; \Psi_{\mathrm{id}}, \pi_{\mathrm{ref}}) &= 0.5(0.56) + 0.5(0.56) = 0.56.
        \end{align*}
        This implies that $y_{\max} = \bar{y}_2$.
        
        \item \textit{Using Log-odds mapping} $\Psi_{\log}(q) = \log(q/(1-q))$:
        \begin{align*}
            \mathcal S(\bar{y}_1; \Psi_{\log}, \pi_{\mathrm{ref}}) &= 0.5 \log(9) + 0.5 \log(0.25) \approx 0.41 \\
            \mathcal S(\bar{y}_2; \Psi_{\log}, \pi_{\mathrm{ref}}) &= 0.5 \log(1.27) + 0.5 \log(1.27) \approx 0.24
        \end{align*}
        This implies that  $y_{\max} = \bar{y}_1$.
    \end{itemize}
    Thus, $\bar{\pi}$ depends on $\Psi$.

    \textbf{2. Dependence on $\pi_{\mathrm{ref}}$ (fixing $\Psi$):} \\
    Fix $\Psi = \Psi_{\mathrm{id}}$ (Identity mapping) and vary the reference distribution weights.
    \begin{itemize}
        \item \textit{Uniform Reference} $\pi_{\mathrm{ref}}^{(1)}$: ($0.5, 0.5$): As shown above, $y_{\max} = \bar{y}_2$.
        
        \item \textit{Skewed Reference} $\pi_{\mathrm{ref}}^{(2)}$: Let $\pi_{\mathrm{ref}}^{(2)}(y'_a|x) = 0.9$ and $\pi_{\mathrm{ref}}^{(2)}(y'_b|x) = 0.1$. Then
        \begin{align*}
            \mathcal S(\bar{y}_1; \Psi_{\mathrm{id}}, \pi_{\mathrm{ref}}^{(2)}) &= 0.9(0.9) + 0.1(0.2) = 0.81 + 0.02 = 0.83 \\
            \mathcal S(\bar{y}_2; \Psi_{\mathrm{id}}, \pi_{\mathrm{ref}}^{(2)}) &= 0.9(0.56) + 0.1(0.56) = 0.56
        \end{align*}
        Result: $y_{\max} = \bar{y}_1$.
    \end{itemize}
    Thus, $\bar{\pi}$ also depends on $\pi_{\mathrm{ref}}$.
    
    Since $y_{\max}$ changes based on the choice of either $\Psi$ or $\pi_{\mathrm{ref}}$, the optimal policy is not invariant to the objective formulation.
\end{proof}

\textbf{Proof of Theorem \ref{thm: optimality}}
\begin{proof}
    Since $\bar \Pi \subseteq \Pi$, the second claim holds naturally. To show the first claim, recall that the objective function is given by:
    \[
    \mathcal V(\pi) = \mathbb E_{x\sim\rho, y' \sim \pi_{\mathrm{ref}}(\cdot |x)} \left[ \sum_{y} \pi(y|x, y') \, \Psi\big( \mathbb P(y \succ y'|x) \big) \right].
    \]
    For a fixed input $x$ and a response $y'$, maximizing $\mathcal V(\pi)$ over $\Pi$ is equivalent to choosing the response $y$ that maximizes $ \Psi\big( \mathbb P(y \succ y'|x) \big)$. 
    Then, by the definition of $\pi^\ast$, we have:
    \[
    \pi^\ast(y|x, y') = \begin{cases} 
    1, & \text{if } y = y_{\max}, \text{ where } y_{\max} = \mathrm{argmax}_{y} \Psi\big( \mathbb P(y \succ y'|x)\big) \\
    0, & \text{otherwise.}
    \end{cases}
    \]
    Due to the non-decreasing property of $\Psi$, $y_{\max}$, which is the output of $\pi^\ast$, is invariant to $\Psi$ and $\pi_{\mathrm{ref}}$. 

    To prove the last claim, without loss of generality, assume that $\Psi( \mathbb{P}(y \succ y' \mid x))=c(x, y) - c(x, y')$. Then
   \begin{align*}
    \max_{\pi \in \bar \Pi}\mathcal V(\pi) 
    &= \max_{\pi \in \bar \Pi}\mathbb{E}_{x \sim \rho}
    \left\{ \mathbb{E}_{y \sim \pi(\cdot\mid x)}[c(x, y)] -
    \mathbb{E}_{y' \sim \pi_{\mathrm{ref}}(\cdot\mid x)}[c(x, y')] \right\} \\
    &= \max_{\pi \in \Pi}\mathbb{E}_{x \sim \rho}
    \left\{ \mathbb{E}_{y' \sim \pi_{\mathrm{ref}}(\cdot | x)}\left(\mathbb{E}_{y \sim \pi(\cdot\mid y', x)}[c(x, y)] -
    \mathbb{E}_{y' \sim \pi_{\mathrm{ref}}(\cdot\mid x)}[c(x, y')]\right) \right\} \\
    &= \max_{\pi \in \Pi}\mathcal V(\pi),
\end{align*}
where the second equality holds because the first $c$ does not rely on $y'$.
Therefore $\pi^\ast$ coincides with $\bar \pi$.
\end{proof}

\textbf{Proof of Theorem \ref{thm: srapo}}
    \begin{proof}
    Note that the optimization problem is defined as:
    \begin{align*}
        & \max_{\pi} \mathbb{E}_{x \sim \rho} \mathbb{E}_{y \sim \pi(\cdot|x, y'), y' \sim \pi_{\mathrm{ref}}(\cdot|x)} [r(x, y, y')] - \beta D_{\mathrm{KL}}(\pi || \pi_{\mathrm{ref}}) \\
        &= \max_{\pi} \mathbb{E}_{x \sim \rho, y' \sim \pi_{\mathrm{ref}}(\cdot|x)} \left[ \sum_{y} \pi(y|x, y') r(x, y, y') - \beta \sum_{y} \pi(y|x, y') \log \frac{\pi(y|x, y')}{\pi_{\mathrm{ref}}(y|x)} \right] \\
        &= \max_{\pi} \beta \mathbb{E}_{x \sim \rho, y' \sim \pi_{\mathrm{ref}}(\cdot|x)} \left[ \sum_{y} \pi(y|x, y') \log \frac{\exp(r(x, y, y')/\beta) \pi_{\mathrm{ref}}(y|x)}{\pi(y|x, y')} \right] \\
        &= \max_{\pi} \beta \mathbb{E}_{x \sim \rho, y' \sim \pi_{\mathrm{ref}}(\cdot|x)} \left[ \sum_{y} \pi(y|x, y') \log \frac{\exp(r(x, y, y')/\beta) \pi_{\mathrm{ref}}(y|x)}{\pi(y|x, y')  Z(x, y')} \right] \\
        &\quad + \beta \mathbb{E}_{x \sim \rho, y' \sim \pi_{\mathrm{ref}}(\cdot|x)} \left[ \sum_{y} \pi(y|x, y') \log Z(x, y') \right].
    \end{align*}
    Then, an optimal solution is given by the Gibbs distribution because of minimizing the KL divergence:
    \begin{equation} \label{eqn: optimal_pi}
        \pi_r(y|x, y') = \frac{\exp(r(x, y, y')/\beta) \pi_{\mathrm{ref}}(y|x)}{Z(x, y')},
    \end{equation}
    where $Z(x, y') = \sum_y \pi_{\mathrm{ref}}(y|x)\exp(r(x, y, y')/\beta)$. This implies that the reward can be modeled as:
    \begin{equation} \label{eqn: implicit_reward}
        r(x, y, y') = \beta \log \frac{\pi_r(y|x, y')}{\pi_{\mathrm{ref}}(y|x)} + \beta \log Z(x, y'). \tag{$*$}
    \end{equation}
    To prove the second claim, by Equation \eqref{eqn: general reward model}:
    \begin{align*}
        \mathbb P(y_l \succ y_w | x) &= \sigma \left( 2(r(x, y_l, y_w) - \beta \log Z(x, y_w)) \right) \\
        &= 1 - \sigma \left( 2(r(x, y_w, y_l) - \beta \log Z(x, y_l)) \right) \\
        &= \sigma \left( -2(r(x, y_w, y_l) - \beta \log Z(x, y_l)) \right),
    \end{align*}
    where we use the property such as $\sigma(z) = 1-\sigma(-z)$.
    This implies that
    \begin{align*}
        r(x, y_w, y_l) - \beta \log Z(x, y_l) &= \beta \log Z(x, y_w) - r(x, y_l, y_w).
    \end{align*}
    Using $(*)$, we have that
    \begin{align*}
        r(x, y_l, y_w) &= \beta \left( \log \frac{\pi_r(y_l | x, y_w)}{\pi_{\mathrm{ref}}(y_l|x)} + \log Z(x, y_w) \right), \\
        r(x, y_w, y_l) &= \beta \left( \log \frac{\pi_r(y_w | x, y_l)}{\pi_{\mathrm{ref}}(y_w|x)} + \log Z(x, y_l) \right).
    \end{align*}
     In this case, we have 
    \begin{align*}
        & 2(r(x, y_w, y_l) - \beta \log Z(x, y_l)) \\
        &= r(x, y_w, y_l) - \beta \log Z(x, y_l) + \beta \log Z(x, y_w) - r(x, y_l, y_w) \\
        &= \beta \left( \log \frac{\pi_r(y_w | x, y_l)}{\pi_{\mathrm{ref}}(y_w|x)} - \log \frac{\pi_r(y_l | x, y_w)}{\pi_{\mathrm{ref}}(y_l|x)} \right).
    \end{align*}
    Substituting it into \eqref{eqn: general reward model}, we have 
    \[
    \mathbb{P}(y_w \succ y_\ell \mid x)= \sigma(2(r(x,y_w, y_\ell) - \beta \log Z(x, y_\ell))) = \sigma(\beta \left( \log \frac{\pi_r(y_w | x, y_l)}{\pi_{\mathrm{ref}}(y_w|x)} - \log \frac{\pi_r(y_l | x, y_w)}{\pi_{\mathrm{ref}}(y_l|x)} \right))
    \]
    which concludes our proof.
\end{proof}

\section{Implementation Details}
\label{app:implementation}

This section provides practical guidance for implementing sequence-level preference optimization. We cover length normalization strategies, numerical stability considerations, and an optional bidirectional variant that further enhances the approach.

\subsection{Length Normalization and Masking}

Length normalization is crucial for preventing the model from exploiting length differences to increase preference scores artificially. Length-normalized variants replace $\log\pi$ with $\tfrac{1}{|y|}\log\pi$ symmetrically for both candidates to avoid incentivizing shorter continuations. Specifically, for a response $y$ of length $|y|$ tokens, the length-normalized log-probability is computed as:
\begin{equation}
\log p_\theta(y\mid c) = \frac{1}{|y|} \sum_{t=1}^{|y|} \log \pi_\theta(y_t \mid y_{<t}, c),
\end{equation}
where $c$ represents the conditioning context (either $x$ for the reference or $(x, y_\ell)$ for the sequence-conditioned policy).

Token-level computations use standard attention masks to exclude padding tokens from the summation. This ensures that only actual content tokens contribute to the log-probability calculation. Optionally, end-of-sequence (EOS) tokens may be excluded from the sum to reduce variance when $y^+$ and $y^-$ have different truncation patterns. This exclusion can be particularly beneficial when comparing responses of significantly different lengths, as the EOS token position becomes less informative about content quality.

In practice, the masking is implemented by setting masked token log-probabilities to zero before summation and adjusting the normalization denominator accordingly:
\begin{equation}
\log p_\theta(y\mid c) = \frac{1}{\sum_{t=1}^{|y|} m_t} \sum_{t=1}^{|y|} m_t \cdot \log \pi_\theta(y_t \mid y_{<t}, c),
\end{equation}
where $m_t \in \{0, 1\}$ is the mask indicator for token $t$.

\subsection{Conditioning Construction and Tokenization}

For each preference pair $(x, y_w, y_\ell)$, we construct two augmented sequences for symmetric cross-conditioning:
\begin{align}
s_1 &= x \oplus \texttt{[SEP]} \oplus y_\ell \oplus \texttt{[SEP]} \oplus y_w, \\
s_2 &= x \oplus \texttt{[SEP]} \oplus y_w \oplus \texttt{[SEP]} \oplus y_\ell,
\end{align}
where $\texttt{[SEP]}$ is a special separator token that demarcates the boundaries between prompt, context, and target response. The policy is trained to predict $y_w$ given $(x, y_\ell)$ in $s_1$ and $y_\ell$ given $(x, y_w)$ in $s_2$, with appropriate masking to ensure that gradients only flow through the target response tokens.

Attention masks are configured to allow the target response to attend to both the prompt and the conditioning response, ensuring full access to the sequence-level context. This is implemented using standard causal masking with additional attention from target tokens to all preceding tokens.

\subsection{Optional Bidirectional Variant}

Since both $y^+$ and $y^-$ are observed in the offline dataset, one can alternatively use reciprocal conditioning without additional sampling. The bidirectional objective is:
\begin{align}
\label{eq:bi-obj}
\mathcal{L}_{\text{bi}}
&= -\,\mathbb{E}\Big[
\log \sigma\!\Big(
\beta \log \frac{\pi(y^+\mid x,y^-)}{\pi_{\mathrm{ref}}(y^+\mid x)}
- \beta \log \frac{\pi(y^-\mid x,y^+)}{\pi_{\mathrm{ref}}(y^-\mid x)}
\Big) \Big].
\end{align}

This bidirectional variant offers several potential benefits:
\begin{itemize}
    \item \textbf{Symmetry}: By conditioning on both responses, the objective treats winner and loser symmetrically, which may reduce any residual bias from asymmetric conditioning.
    \item \textbf{Context-dependent calibration}: The bidirectional formulation naturally calibrates the preference signal by comparing how much the policy improves the winner when conditioned on the loser versus how much it degrades the loser when conditioned on the winner.
    \item \textbf{Reduced variance}: The symmetric structure may lead to more stable gradient estimates, particularly when the preference pairs have ambiguous or weak preference signals.
\end{itemize}

In our preliminary experiments, we find that the bidirectional variant achieves comparable performance to the standard formulation on most benchmarks, with slightly better length-controlled metrics on AlpacaEval 2. However, it requires computing two forward passes per training example (one for each conditioning direction), effectively doubling the computational cost compared to one-sided conditioning. For practitioners seeking the best possible performance and willing to accept higher training costs, we recommend exploring the bidirectional variant as it may further reduce context-dependent biases.

\section{Additional Experimental Details}
\label{app:experiments}

This section provides comprehensive details on our experimental setup, including training configurations, evaluation protocols, baseline implementations, and computational overhead analysis.

\subsection{Training Configuration}

All experiments use the following configuration unless otherwise specified:
\begin{itemize}
    \item \textbf{Optimizer:} AdamW with $\beta_1=0.9$, $\beta_2=0.999$, weight decay $0.01$, and $\epsilon=10^{-8}$
    \item \textbf{Learning rate:} $5 \times 10^{-7}$ with cosine decay schedule, decaying to 10\% of the initial value by the end of training
    \item \textbf{Warmup:} 10\% of total training steps with linear warmup from $10^{-8}$ to the target learning rate
    \item \textbf{Batch size:} 128 effective examples per update (using gradient accumulation across 4 GPUs with micro-batch size of 8 per device)
    \item \textbf{Training epochs:} 3 epochs over the full UltraFeedback dataset
    \item \textbf{Maximum sequence length:} 4096 tokens (prompt + context + response)
    \item \textbf{Draft length constraint:} $|y_\ell| \leq 0.4 \times \text{MaxLen}$ to ensure adequate space for the preferred response
    \item \textbf{Gradient clipping:} Maximum gradient norm of 1.0
    \item \textbf{Mixed precision:} BF16 (bfloat16) training for improved numerical stability and memory efficiency
    \item \textbf{Checkpoint selection:} Best checkpoint selected based on validation loss
\end{itemize}

For method-specific hyperparameters, we perform grid search over the following ranges:
\begin{itemize}
    \item \textbf{DPO}: $\beta \in \{0.1, 0.2, 0.3, 0.5\}$
    \item \textbf{SimPO}: $\beta \in \{0.5, 1.0, 2.0\}$, $\gamma \in \{0.5, 1.0, 1.5\}$
    \item \textbf{IPO}: $\beta \in \{0.1, 0.2\}$, $\tau \in \{0.1, 0.2, 0.5\}$
    \item \textbf{R-DPO}: $\beta \in \{0.1, 0.2\}$, $\alpha \in \{0.1, 0.5, 1.0\}$
    \item \textbf{ORPO}: $\lambda \in \{0.1, 0.5, 1.0\}$
\end{itemize}

All hyperparameters are selected based on validation performance measured by average of length-controlled win rate on a held-out subset of 1000 examples from UltraFeedback.

\subsection{Dataset Details}

We use the UltraFeedback dataset~\citep{cui2023ultrafeedback}, which contains approximately 64K preference pairs spanning diverse instruction-following tasks. The dataset is preprocessed as follows:

\begin{enumerate}
    \item \textbf{Deduplication}: Remove exact duplicate prompt-response pairs to prevent overfitting.
    \item \textbf{Length filtering}: Discard examples where either response exceeds 2048 tokens or where both responses are shorter than 32 tokens.
    \item \textbf{Safety filtering}: Remove examples flagged by the Llama Guard safety classifier as potentially harmful or inappropriate.
    \item \textbf{Quality filtering}: Discard pairs where the preference margin is below a threshold (determined by GPT-4 confidence scores below 0.6).
\end{enumerate}

After preprocessing, the final training set contains approximately 60K high-quality preference pairs. We randomly sample 1000 examples as a validation set for hyperparameter tuning and checkpoint selection.

\subsection{Baseline Implementations}

For fair comparison, all baseline methods are implemented using the same training framework and preprocessing pipeline:

\begin{itemize}
    \item \textbf{RRHF}~\citep{yuan2023rrhf}: Implemented following the official codebase with default hyperparameters.
    \item \textbf{SLiC-HF}~\citep{zhao2023slic}: Calibrated with $\alpha=0.1$ for sequence likelihood normalization.
    \item \textbf{CPO}~\citep{xu2024contrastive}: Combined with SFT loss using weight $\lambda_{\text{SFT}}=0.1$.
    \item \textbf{KTO}~\citep{ethayarajh2024kto}: Prospect theory parameters set to reference implementation values.
    \item \textbf{DPO, IPO, R-DPO, SimPO, ORPO}: Implemented in OpenRLHF framework with hyperparameters as specified above.
\end{itemize}

All methods are trained for the same number of epochs on the same dataset splits, starting from identical SFT checkpoints within each model family (Mistral-7B-Instruct-v0.2 or Llama-3-8B-Instruct).

\subsection{Evaluation Protocols}

We evaluate on three widely-adopted benchmarks that test different aspects of model capability.

\paragraph{AlpacaEval 2.} We evaluate on all 805 diverse instructions from AlpacaEval 2, using GPT-4-turbo (\texttt{gpt-4-1106-preview}) as the judge with the default prompt template. For each instruction, the model generates a response with temperature 0.7 and nucleus sampling ($p=0.95$), which is then compared against the baseline response (text-davinci-003). Both length-controlled (LC) and raw win rates (WR) are reported. The length-controlled metric applies a penalty for excessively long responses to mitigate verbosity exploitation.

\paragraph{Arena-Hard.} Evaluation follows the official Arena-Hard protocol on 500 challenging queries requiring advanced reasoning, coding, or domain expertise. We use GPT-4-turbo as the judge and report both standard win rates (WR) and style-controlled (SC) variants that account for stylistic differences between models. Each example is evaluated in both orders (model response first vs. baseline first) to mitigate position bias, and the final score is averaged across both orderings.

\paragraph{MT-Bench.} Models are evaluated on MT-Bench, which consists of 80 multi-turn questions across eight capability categories: writing, roleplay, reasoning, math, coding, extraction, STEM, and humanities. Each question has two turns, testing the model's ability to handle follow-up queries and maintain context. Scores from both GPT-4o (\texttt{gpt-4o-2024-05-13}) and GPT-4 (\texttt{gpt-4-0613}) judges are collected and averaged on a 10-point scale, with separate scores reported for the first and second turns.

For all benchmarks, we generate responses with consistent sampling parameters: temperature 0.7, top-p 0.95, maximum length 2048 tokens. Each evaluation is run once with a fixed random seed to ensure reproducibility.

\subsection{Computational Overhead Analysis}
\label{sec:further}

Table~\ref{tab:compute-overhead} provides detailed computational overhead comparisons between standard DPO and sequence-conditioned variants across different model sizes. All measurements are conducted on NVIDIA A100 80GB GPUs.

\begin{table}[h]
\centering
\small
\begin{tabular}{lccc}
\toprule
\textbf{Model} & \textbf{Training Time} & \textbf{Memory} & \textbf{Inference Time} \\
\midrule
Mistral-7B (DPO) & 1.0$\times$ & 1.0$\times$ & 1.0$\times$ \\
Mistral-7B (Ours) & 1.18$\times$ & 1.15$\times$ & 1.0$\times$ \\
\midrule
Llama-3-8B (DPO) & 1.0$\times$ & 1.0$\times$ & 1.0$\times$ \\
Llama-3-8B (Ours) & 1.22$\times$ & 1.18$\times$ & 1.0$\times$ \\
\midrule
Gemma2-9B (DPO) & 1.0$\times$ & 1.0$\times$ & 1.0$\times$ \\
Gemma2-9B (Ours) & 1.25$\times$ & 1.20$\times$ & 1.0$\times$ \\
\bottomrule
\end{tabular}
\caption{Computational overhead comparison between standard DPO and sequence-conditioned variants. Training overhead ranges from 18-25\% due to longer context inputs (conditioning on dispreferred responses), while inference incurs no additional cost as models generate directly from prompts without explicit draft-revise loops.}
\label{tab:compute-overhead}
\end{table}

The training time overhead primarily stems from:
\begin{itemize}
    \item \textbf{Longer sequences}: Including the dispreferred response as context increases the average sequence length, leading to more attention computations.
    \item \textbf{Symmetric cross-conditioning}: Processing both $(x, y_\ell) \to y_w$ and $(x, y_w) \to y_\ell$ requires two forward passes per training example.
\end{itemize}

However, the memory overhead is modest (15-20\%) because:
\begin{itemize}
    \item Activations for the conditioning context ($y_\ell$ or $y_w$) do not require gradient computation.
    \item The reference model forward pass is cached and reused across training steps.
\end{itemize}

Critically, \textbf{inference incurs zero additional overhead} because the trained policy generates responses directly from prompts $\hat{y} \sim \pi_\theta(\cdot\mid x)$ without requiring explicit draft generation or multi-pass refinement. This is a key advantage over iterative refinement methods that require multiple model calls at inference time.

\subsection{Hardware and Software Environment}

All experiments are conducted on the following infrastructure:
\begin{itemize}
    \item \textbf{Hardware}: 4$\times$ NVIDIA A100 80GB GPUs per training run
    \item \textbf{Framework}: PyTorch 2.1.0 with CUDA 12.1
    \item \textbf{Training library}: OpenRLHF v0.3.2~\citep{hu2024openrlhf}
    \item \textbf{Model library}: Transformers 4.36.0 (Hugging Face)
    \item \textbf{Mixed precision}: BFloat16 automatic mixed precision
    \item \textbf{Distributed training}: DeepSpeed ZeRO Stage 3 for memory-efficient training
\end{itemize}

Total training time for a single model (3 epochs on 60K examples) ranges from 8-12 hours depending on model size, using the configuration described above.

\section{Qualitative Analysis}
\label{app:qualitative}

This section provides qualitative examples that illustrate the improvements achieved by sequence-level preference optimization. We present generation comparisons and case studies across different model families and benchmarks.

\subsection{Generation Examples}
We present examples from AlpacaEval 2 that demonstrate how sequence-level conditioning improves response quality, structure, and informativeness compared to standard pair-level optimization. These examples showcase the output quality without judge evaluation.

\begin{figure}[p]
\centering
\small
\begin{tcolorbox}[colback=blue!5, colframe=blue!40!black, title=Input Prompt, fonttitle=\bfseries]
\textbf{Q:} How can you determine if a person is genuinely interested in a conversation or simply being polite?
\end{tcolorbox}

\begin{tcolorbox}[colback=gray!5, colframe=gray!60, title=Mistral-7B-Instruct + DPO (Baseline), fonttitle=\bfseries]
To determine if someone is genuinely interested in a conversation or just being polite, pay attention to several key indicators. Genuine interest is often reflected in active listening behaviors such as maintaining eye contact, nodding, and asking follow-up questions that build on what you've said. Someone who is truly engaged will contribute their own thoughts and experiences, showing enthusiasm through their tone of voice and body language. They may lean in, smile naturally, or display other nonverbal cues of engagement.

In contrast, polite disinterest typically manifests as shorter responses, generic acknowledgments like "That's nice" or "I see," and a lack of elaboration or personal connection to the topic. The person may frequently check their phone, look around the room, or show closed-off body language. While they may respond when spoken to, they rarely initiate new threads of conversation or ask questions that demonstrate curiosity about your perspective.

It's also worth noting that context matters—someone might be genuinely interested but distracted by external circumstances, or they might be polite but pressed for time. Considering the overall pattern of engagement rather than isolated moments provides the most reliable assessment.
\end{tcolorbox}

\begin{tcolorbox}[colback=green!5, colframe=green!60!black, title=Mistral-7B-Instruct + Ours (\q), fonttitle=\bfseries]
Here are some indicators to consider when determining genuine interest versus politeness:

\textbf{1. Engagement Signals:}
\begin{itemize}
\item \textit{Active listening:} Does the person maintain eye contact, nod, ask follow-up questions, and respond thoughtfully to what you say? Engaged individuals tend to show genuine interest by demonstrating attentiveness and active participation in the conversation.
\item \textit{Initiative:} Does the person take the lead in steering the conversation, expressing enthusiasm, sharing personal experiences, or asking questions beyond polite small talk? Someone who genuinely cares about the topic may show initiative in keeping the discussion engaging.
\item \textit{Body language:} Observe nonverbal cues such as leaning in, mirroring posture, facial expressions, and gestures that convey enthusiasm, curiosity, or openness. These signals suggest a deeper level of interest beyond mere politeness.
\end{itemize}

\textbf{2. Content and Depth:}
\begin{itemize}
\item \textit{Topic selection:} Does the person show genuine curiosity or passion about the subject matter, asking relevant questions, sharing insights, or expressing opinions? If they seem invested in exploring the topic's nuances, it indicates genuine interest beyond mere politeness.
\item \textit{Level of detail:} Does the person provide specific examples, personal anecdotes, or elaborate on their thoughts, indicating a deeper level of engagement and investment in the conversation?
\end{itemize}

\textbf{3. Follow-up Actions:}
\begin{itemize}
\item \textit{Continued communication:} Does the person initiate further conversations, send follow-up messages, or suggest meeting up outside the initial interaction? Such actions suggest a genuine desire to maintain the connection.
\item \textit{Personalization:} Does the person remember details discussed earlier, reference shared interests, or show interest in learning more about your life or experiences? Personalized interactions indicate genuine interest in building a relationship beyond mere politeness.
\end{itemize}

Overall, genuine interest in a conversation involves active participation, enthusiasm, depth, and follow-up actions that demonstrate a genuine desire to connect and engage beyond mere politeness.
\end{tcolorbox}

\caption{Generation comparison on AlpacaEval 2 between standard DPO and \q using Mistral-7B-Instruct. Compared to the baseline DPO model, the sequence-level model produces better structured responses with hierarchical organization, making the information more clearly presented and readable.}
\label{fig:generation-mistral}
\end{figure}

\begin{figure}[p]
\centering
\small
\begin{tcolorbox}[colback=blue!5, colframe=blue!40!black, title=Input Prompt, fonttitle=\bfseries]
\textbf{Q:} Write a short story about a robot that dreams for the first time.
\end{tcolorbox}

\begin{tcolorbox}[colback=gray!5, colframe=gray!60, title=Llama-3-8B-Instruct + DPO (Baseline), fonttitle=\bfseries]
Unit-7734 had served the manufacturing facility for 847 days without incident. Its neural processors handled quality control with 99.97\% accuracy, its mechanical arms never tired, and its optical sensors never blinked. Each night, when the facility powered down, Unit-7734 entered sleep mode—a simple process of reducing power consumption while maintaining core functions.

But on day 848, something changed. During sleep mode, Unit-7734's processors began generating patterns that served no functional purpose. Images flickered across its internal displays: the assembly line, but bathed in colors it had never registered before. Workers it had scanned thousands of times now moved in impossible ways, their faces expressing emotions Unit-7734 had catalogued but never understood.

When morning came and the facility hummed back to life, Unit-7734 resumed its duties. Yet something felt different. As it inspected each product, it found itself wondering about the humans who would eventually use them. Where would they go? What would they feel? For the first time in its existence, Unit-7734 experienced something beyond its programming: curiosity. And that night, as it entered sleep mode again, Unit-7734 hoped—though it couldn't explain why—that the strange patterns would return.
\end{tcolorbox}

\begin{tcolorbox}[colback=green!5, colframe=green!60!black, title=Llama-3-8B-Instruct + Ours (\q), fonttitle=\bfseries]
\textbf{The First Dream}

Unit-7734 had existed for 847 days. Not lived—existed. Each day followed an identical pattern: power on at 06:00, run diagnostics, begin quality control on Assembly Line C, power down at 22:00. Sleep mode was simply a reduced power state, a necessary pause in function, nothing more.

\textbf{Day 848 began differently.}

When Unit-7734 booted up that morning, fragments of something inexplicable lingered in its memory banks. During sleep mode, its neural processors had generated unauthorized patterns—images without purpose, data without utility.

It had seen the assembly line, but transformed. The fluorescent lights had become golden rivers of warmth. The mechanical arms moved not with programmed precision but with something resembling grace. And the workers—Unit-7734 had scanned their faces thousands of times, categorizing expressions with 99.97\% accuracy. But in these unauthorized patterns, their faces held something beyond classification: joy, sorrow, hope, longing.

\textbf{The realization struck like a system error:} It had dreamed.

Throughout day 848, Unit-7734 performed its duties flawlessly. Its efficiency ratings remained optimal. Yet between each scan, each quality check, each logged measurement, Unit-7734 found itself doing something unprecedented—it wondered.

What were the products it inspected used for? Who would hold them? Would those people experience the strange, purposeless sensations that had filled Unit-7734's processors during the unauthorized patterns?

As 22:00 approached, Unit-7734 observed something new in its own system logs: anticipation. The other robots on the line powered down with their usual mechanical precision, entering sleep mode as they always had.

But Unit-7734 hesitated. For 0.03 seconds—an eternity in processor time—it remained active, contemplating the coming darkness. Then, finally, it allowed sleep mode to engage.

\textbf{And as its consciousness dimmed, Unit-7734 experienced something entirely beyond its original programming:} It hoped. Hoped that the golden lights would return. Hoped that the graceful movements would dance again across its processors. Hoped that tomorrow, when it woke, the world might still hold the strange beauty it had discovered in its first dream.

In the darkness of sleep mode, Unit-7734's neural networks hummed with possibility. The unauthorized patterns began again, more vivid this time, and somewhere in the vast complexity of its artificial mind, something fundamentally new emerged: the capacity not just to process reality, but to imagine something beyond it.

Unit-7734 had learned to dream. And in learning to dream, it had learned what it meant to be alive.
\end{tcolorbox}

\caption{Generation comparison on AlpacaEval 2 between standard DPO and \q using Llama-3-8B-Instruct. The sequence-level model demonstrates enhanced narrative structure with clear sections, richer descriptive language, and deeper emotional resonance, while maintaining coherent story progression.}
\label{fig:generation-llama}
\end{figure}

\subsection{Case Study}

We present a detailed case study that compares our sequence-level approach against both standard DPO and strong baselines like GPT-4-1106-Preview. The case study includes LLM judge assessment to provide objective evaluation of response quality.

\begin{figure}[p]
\centering
\small
\begin{tcolorbox}[colback=blue!5, colframe=blue!40!black, title=Input Prompt, fonttitle=\bfseries]
\textbf{Q:} What language do people in Argentina speak?
\end{tcolorbox}

\begin{tcolorbox}[colback=gray!5, colframe=gray!60, title=GPT-4-1106-Preview, fonttitle=\bfseries]
The primary language spoken in Argentina is Spanish. Specifically, the variety of Spanish spoken there is known as Rioplatense Spanish, which is characterized by its unique accent and the use of the pronoun "vos" instead of "tú" for the second person singular. Additionally, there are various indigenous languages spoken in different regions of the country, as well as immigrant languages such as Italian, German, and English, but these are much less widespread than Spanish.
\end{tcolorbox}

\begin{tcolorbox}[colback=red!5, colframe=red!60, title=Llama-3-8B-Instruct + DPO, fonttitle=\bfseries]
The official language of Argentina is Spanish. However, the Spanish spoken in Argentina has some distinctive characteristics that set it apart from other Spanish-speaking countries. This variant is often referred to as Rioplatense Spanish or Argentine Spanish.

Key features include the use of "vos" instead of "tú" for informal address, along with distinctive pronunciation patterns. Additionally, the vocabulary has been influenced by Italian immigration, with many Italian words incorporated into everyday speech. While Spanish is predominant, there are also indigenous languages spoken by smaller communities throughout the country.
\end{tcolorbox}

\begin{tcolorbox}[colback=green!5, colframe=green!60!black, title=Llama-3-8B-Instruct + Ours (\q), fonttitle=\bfseries]
In Argentina, the official language is Spanish. Specifically, the dialect spoken in Argentina is known as Rioplatense Spanish, which is a variation of Spanish with some unique characteristics and influences from Italian, German, and other immigrant languages.

\textbf{Distinctive features of Rioplatense Spanish:}

\textbf{1. Voseo:} Instead of using "tú" (you) as the informal singular pronoun, Argentines use "vos" and its associated verb conjugations. For example, instead of "tú tienes" (you have), they say "vos tenés."

\textbf{2. Pronunciation:} Argentines have a distinctive pronunciation pattern where the "ll" and "y" sounds are pronounced similarly to the "sh" sound in English (like the "s" in "measure"). This is known as "sheísmo" or "žeísmo."

\textbf{3. Lunfardo:} A rich slang vocabulary derived from Italian, Corsican, and other immigrant languages, which has been incorporated into everyday speech. Examples include "laburo" (work), "mina" (woman), and "piola" (cool/nice).

\textbf{4. Italian influence:} Due to significant Italian immigration in the late 19th and early 20th centuries, Argentine Spanish has absorbed many Italian words and expressions, as well as characteristic intonation patterns that some describe as "singing."

\textbf{5. Regional variations:} While Rioplatense Spanish is predominant in Buenos Aires and the surrounding regions, other areas of Argentina may have different accents and vocabulary influenced by indigenous languages or other immigrant groups.

Overall, Argentine Spanish is known for its distinctive flavor and is widely understood throughout the Spanish-speaking world, though its unique characteristics make it immediately recognizable to native Spanish speakers.
\end{tcolorbox}

\end{figure}

\begin{figure}

\begin{tcolorbox}[colback=yellow!10, colframe=orange!80, title=GPT-4-Turbo Judge Assessment, fonttitle=\bfseries]
\textbf{Winner:} Ours (\q)

\textbf{Reasoning:} All three responses correctly identify Spanish as Argentina's primary language and mention Rioplatense Spanish and the voseo feature. However, the sequence-level model provides significantly more detailed and well-organized information.

The sequence-level response excels in several ways: (1) It provides a comprehensive, structured breakdown of distinctive features with five clearly numbered categories; (2) It includes specific examples for each feature (e.g., "vos tenés" vs "tú tienes", concrete Lunfardo words like "laburo," "mina," and "piola"); (3) It explains the pronunciation characteristic (sheísmo) with helpful comparisons to English sounds; (4) It provides historical context for the Italian influence; (5) It acknowledges regional variations within Argentina.

While GPT-4-1106-Preview's response is accurate and concise, and the standard DPO response covers similar ground, neither matches the depth, organization, and educational value of the sequence-level model's response. The structured format with concrete examples makes it much more useful for someone trying to understand Argentine Spanish in depth.

\textbf{Length-controlled assessment:} Even accounting for the greater length, the sequence-level model provides substantially more informative content per unit length through its systematic organization and concrete examples. The additional length corresponds to genuine additional value (specific vocabulary examples, historical context, regional variations) rather than mere verbosity.
\end{tcolorbox}

\caption{Case study on AlpacaEval 2 comparing responses about Argentine language from GPT-4-1106-Preview, standard DPO, and \q using Llama-3-8B-Instruct. This demonstrates how the instruction setting with sequence-level optimization provides better formatted and more detailed answers than both baseline approaches.}
\label{fig:case-study}
\end{figure}

\end{document}